\documentclass[11pt]{article}
\pdfoutput=1
\usepackage{enumerate}
\usepackage{pdfsync}
\usepackage{times}
\usepackage[OT1]{fontenc}
\usepackage[usenames]{color}
\usepackage{booktabs}
\usepackage{smile}
\usepackage[colorlinks,
            linkcolor=red,
            anchorcolor=black,
            citecolor=black
            ]{hyperref}
\usepackage{mathrsfs}
\usepackage{fullpage}
\usepackage{hyperref}
\usepackage[protrusion=true,expansion=true]{microtype}
\usepackage{float}
\usepackage{subfigure}
\usepackage{setspace}
\usepackage{algorithm}
\usepackage{algorithmic}
\usepackage{balance}
\usepackage{enumitem}

\usepackage{color}

\makeatletter
\newcommand*{\rom}[1]{\expandafter\@slowromancap\romannumeral #1@}
\makeatother

\newcommand{\shortsection}[1]{\vspace{1ex}\noindent{\bf #1.}}

\begin{document}
\title{\huge Efficient Privacy-Preserving Stochastic Nonconvex Optimization}
\author{
 Lingxiao Wang$^*$, Bargav Jayaraman$^\dag$, David Evans$^\dag$, Quanquan Gu$^\S$ \\[1.5ex]
$*$: \emph{Toyota Technological Institute at Chicago} \\
 {\small \sf lingxw@ttic.edu} \\[1.2ex]
 $\dag$: \emph{Department of Computer Science, University of Virginia} \\
 {\small \sf [bj4nq, evans]@virginia.edu} \\[1.2ex]
  $\S$: \emph{Department of Computer Science, University of California, Los Angeles} \\
 {\small \sf qgu@cs.ucla.edu} 
}

\date{}
\maketitle

\begin{abstract} 
While many solutions for privacy-preserving convex empirical risk minimization (ERM) have been developed, privacy-preserving nonconvex ERM remains a challenge. We study nonconvex ERM, which takes the form of minimizing a finite-sum of nonconvex loss functions over a training set. We propose a new differentially private stochastic gradient descent algorithm for nonconvex ERM that achieves strong privacy guarantees efficiently, and provide a tight analysis of its privacy and utility guarantees, as well as its gradient complexity. Our algorithm reduces gradient complexity while improves the best previous utility guarantee given by Wang et al.\ (NeurIPS 2017). Our experiments on benchmark nonconvex ERM problems demonstrate superior performance in terms of both training cost and utility gains compared with previous differentially private methods using the same privacy budgets.
\end{abstract}

\section{Introduction}\label{sec:intro}
For many important domains such as health care and medical research, the datasets used to train machine learning models contain sensitive personal information. There is a risk that models trained on this data can reveal private information about individual records in that training data~\citep{fredrikson2014privacy,shokri2017membership,carlini2018secret}. 
This motivates the research on privacy-preserving machine learning, much of which has focused on achieving \emph{differential privacy}~\citep{dwork2006calibrating}, a rigorous definition of privacy that provides statistical data privacy for individual records. In the past decade, many differentially private machine learning algorithms for solving the empirical risk minimization (ERM) problem 
have been proposed (e.g., \citep{chaudhuri2011differentially,kifer2012private,bassily2014private,zhang2017efficient,wang2017differentially,jayaraman2018distributed,wang2019differentiallyhd,wang2019knowledge}). Almost all of these are for ERM with convex loss functions, 
but many important machine learning approaches, including deep learning, are formulated as ERM problems with nonconvex loss functions.
Furthermore, these learning problems often require large training sets, necessitating the use of stochastic optimization algorithms such as stochastic gradient descent (SGD).

Several recent studies have advanced the application of differential privacy in deep learning~\citep{abadi2016deep,papernot2016semi,brendan2018learning,bu2019deep}. The studies prove differential privacy is satisfied, but evaluate utility experimentally. Only a few differentially private algorithms for solving nonconvex optimization problems have proven utility bounds \citep{zhang2017efficient,wang2017differentially}. For example, \citet{wang2017differentially} proposed a differentially private gradient descent (DP-GD) algorithm
with both privacy and utility guarantees. However, each iteration of DP-GD requires computing the full gradient, which makes it too expensive for use on large training sets. \citet{zhang2017efficient} proposed a random round private stochastic gradient descent (RRPSGD) that can achieve the same privacy guarantee as DP-GD with reduced runtime complexity, but with slightly worse utility bounds. 
In this paper, we propose a differentially private Stochastic Recursive Momentum (DP-SRM) algorithm for nonconvex ERM. 
At the core of our algorithm is the stochastic recursive momentum technique \citep{cutkosky2019momentum} that can consistently reduce the accumulated variance of the gradient estimator. Our approach is more scalable than stochastic variance reduced algorithms \citep{johnson2013accelerating,reddi2016stochastic,allen2016variance,lei2017non,nguyen2017sarah,fang2018spider,zhou2018stochastic} since it eliminates the periodical computation of the checkpoint gradient which usually requires a giant batch size.

\shortsection{Contributions} The main contributions of our paper are summarized as follows:
\begin{itemize}
    \item We develop a new differentially private stochastic optimization algorithm for nonconvex ERM and provide a sharp analysis of the privacy guarantee using R\'enyi Differential Privacy (RDP) \citep{mironov2017renyi}.
    \item Our algorithm improves the best-known utility guarantee for nonconvex optimization, with lower computational complexity. The utility guarantee of our algorithm is $O\big((d\log(1/\delta))^{1/3}/(n\epsilon)^{2/3}\big)$, which is better than the best-known results $O\big((d\log(1/\delta))^{1/4}/(n\epsilon)^{1/2}\big)$ established in \citet{wang2017differentially}. The gradient complexity (i.e., the number of stochastic gradients calculated in total) of our algorithm is $O\big((n\epsilon)^{2}/(d\log(1/\delta))\big)$, which outperforms the best previous results \citep{zhang2017efficient,wang2017differentially} when the problem dimension $d$ is large (see Table \ref{table:Comparision} for more details).
    \item We evaluate our proposed methods on two nonconvex ERM techniques: nonconvex logistic regression and convolutional neural networks.  We report on experiments on several benchmark datasets (Section~\ref{sec:exp}), finding that our method not only produces models that are the closest to the non-private models in terms of model accuracy but also reduces the computational cost.
\end{itemize}

\noindent\textbf{Notation.}
We use curly symbol such as $\cB$ to denote the index set. For a set $\cB$, we use $|\cB|$ to denote its cardinality. For a finite sum function $F=\sum_{i=1}^nf_i/n$, we denote $F_{\cB}$ by $\sum_{i\in\cB}f_i/|\cB|$. For a $d$-dimensional vector $\xb\in\RR^d$, we use $\|\xb\|_2$ to denote its $\ell_2$-norm. Given two sequences $\{a_n\}$ and $\{b_n\}$, if there exists a constant $0<C<\infty$ such that $a_n\leq Cb_n$, we write $a_n = O(b_n)$. Besides, if there exist constants $0<C_1,C_2<\infty$ such that $C_1b_n\leq a_n\leq C_2b_n$, we write $a_n=\Theta(b_n)$. We use $n$, $d$ to represent the number of training examples and the problem dimension, respectively. We also use the standard notation for $(\epsilon,\delta)$-DP where $\epsilon$ is the privacy budget and $\delta$ is the failure probability.

\section{Related Work}\label{sec:related}
Over the past decade, many differentially private machine learning algorithms for convex ERM have been proposed. There are three main approaches to achieve differential privacy in such settings, including output perturbation~\citep{wu2017bolt,zhang2017efficient}, objective perturbation~\citep{chaudhuri2011differentially,kifer2012private,iyengartowards}, and gradient perturbation~\citep{bassily2014private,wang2017differentially,jayaraman2018distributed}. However, other than the methods using gradient perturbation, it is very hard to generalize these methods to nonconvex ERM because of the difficulty in computing the sensitivity for nonconvex ERM. 
Thus, most differentially private algorithms for nonconvex ERM are based on the gradient perturbation, including our work. The problem with gradient perturbation approaches is that their iterative nature quickly consumes any reasonable privacy budget. Hence, the main challenge is to develop algorithms for nonconvex ERM that can provide sufficient utility while maintaining privacy with high computational efficiency.

Several recent works~\citep{abadi2016deep,papernot2016semi,xie2018differentially} studied deep learning with differential privacy. \citet{abadi2016deep} proposed a method called moments accountant to keep track of the privacy cost of stochastic gradient descent algorithm during the training process, which provides a strong privacy guarantee.
\citet{papernot2016semi} established a Private Aggregation of Teacher Ensembles (PATE) framework to improve the privacy guarantee of deep learning for classification tasks.  \citet{xie2018differentially} and \citet{yoon2018pate} investigated the differentially private Generative Adversarial Nets (GAN) with different distance metrics. However, none of these works provide utility guarantees for their algorithms.


\begin{table*}[tb]

    \caption{Comparison of different $(\epsilon,\delta)$-DP algorithms for nonconvex optimization. We report the utility bound in terms of $\EE\|\nabla F(\btheta^p)\|_2$, where $\btheta^p$ is the output of the differentially private algorithm, $\EE$ is taken over the randomness of the algorithm. We only present results in terms of $n,d,\epsilon,\delta$ and ignores other parameters for simplicity.}
	\begin{center}
		\begin{tabular}{cccccc}
			\toprule
			Algorithm  & Utility &Gradient Complexity\\
			\midrule
						RRPSGD \citep{zhang2017efficient} & $O\Big(\frac{(d\log(n/\delta)\log(1/\delta))^{1/4}}{(n\epsilon)^{1/2}}\Big)$&$O\big(n^2\big)$\\
			\midrule
			DP-GD \citep{wang2017differentially}&$O\Big(\frac{(d\log(1/\delta))^{1/4}}{(n\epsilon)^{1/2}}\Big)$&$O\Big(\frac{n^2\epsilon}{(d\log(1/\delta))^{1/2}}\Big)$\\
			\midrule

			DP-SRM &\multirow{2}{*}{$O\Big(\frac{(d\log(1/\delta))^{1/3}}{(n\epsilon)^{2/3}}\Big)$}& \multirow{2}{*}{$O\Big(\frac{(n\epsilon)^{2}}{d\log(1/\delta)}\Big)$}\\(This paper)&&&&&\\
			\bottomrule
		\end{tabular}
		\label{table:Comparision}
	\end{center}
\end{table*}
Table \ref{table:Comparision} summarizes differentially private nonconvex optimization algorithms 
that provide utility guarantees for nonconvex ERM. The Random Round Private Stochastic Gradient Descent (RRPSGD) method developed by \citet{zhang2017efficient} is the first differentially private nonconvex optimization algorithm with the utility guarantee. 
This method performs the perturbed SGD (adding Gaussian noise to the stochastic gradients), for a random number of iterations \citep{ghadimi2013stochastic}. The gradient complexity of RRPSGD is $O(n^2)$, which makes it impractical for most settings. 
\citet{zhang2017efficient} showed that RRPSGD is able to find a stationary point in expectation with a diminishing error $O\big((d\log(n/\delta)\log(1/\delta))^{1/4}/(n\epsilon)^{1/2}\big)$. Their analysis of the privacy guarantee is based on the standard privacy-amplification 
by subsampling result and strong composition theorem \citep{bassily2014private}. Although such an analysis can be easily adapted to the nonconvex setting with stochastic optimization algorithms, it results in a large bound on the variance of the added noise compared with relaxed definitions such as the moments accountant~\citep{abadi2016deep} and Gaussian differential privacy \citep{dong2019gaussian}. 

\citet{wang2017differentially} proposed the Differentially Private Gradient Descent (DP-GD) algorithm for nonconvex optimization. DP-GD has a comparable gradient complexity $O\big(n^2\epsilon/d^{1/2}\big)$, and an improved utility guarantee $O\big((d\log(1/\delta))^{1/4}/(n\epsilon)^{1/2}\big)$ compared with that of RRPSGD. The reason DP-GD can achieve this factor of $O\big((\log(n/\delta))^{1/4}\big)$ improvement, is that it uses the full gradient rather than the stochastic gradient. This makes DP-GD computationally very expensive or even intractable for large-scale machine learning problems ($n$ is big). Recently, \citet{wang2019differentially} also proposed a differentially private stochastic algorithm for nonconvex optimization. 
Their goal is to find the local minima, while we aim to find the stationary point. In addition, their utility guarantee is asymptotic---it provides the desired utility guarantee only if an infinite number of iterations could be run. In contrast, our utility guarantee holds for a finite number of iterations.

\section{Preliminaries}\label{sec:preliminary}
We consider the empirical risk minimization (ERM) problem: given a training set $S=\{(\xb_1,y_1),\ldots,(\xb_n,y_n)\}$ drawn from some unknown but fixed data distribution with $\xb_i \in \RR^D, y_i \in \cY \subseteq \RR$, we aim to find a solution $\hat \btheta \in \RR^d$ that minimizes the following empirical risk
\begin{align}\label{eq:finite_sum}
F(\btheta):=\frac{1}{n}\sum_{i=1}^nf_i(\btheta),
\end{align}
where $F(\btheta)$ is the empirical risk function (i.e., training loss), $f_i(\btheta) = \ell(\btheta;\xb_i,y_i)$ is the loss function defined on the $i$-th training example $(\xb_i,y_i)$, and $\btheta \in \RR^d$ is the model parameter we want to learn.

Here, we provide some definitions and lemmas that will be used in our theoretical analysis. 

\begin{definition}
    $\btheta\in \RR^d$ is an $\zeta$-approximate stationary point if $\|\nabla f(\btheta)\|_2\leq \zeta$.
\end{definition}

\begin{definition}
	A function $f:\RR^d\rightarrow\RR$ is $G$-Lipschitz, if for all $\btheta_1,\btheta_2\in\RR^d$, we have 
	\begin{align*}
	    |f(\btheta_1)- f(\btheta_2)|\leq G\|\btheta_1-\btheta_2\|_2.
	\end{align*}
\end{definition}

\begin{definition}
	A function $f:\RR^d\rightarrow\RR$ has $L$-Lipschitz gradient, if for all $\btheta_1,\btheta_2\in\RR^d$, we have
	\begin{align*}
	\|\nabla f(\btheta_1)- \nabla f(\btheta_2)\|_2\leq L\|\btheta_1-\btheta_2\|_2.
	\end{align*}
\end{definition}



Differential privacy provides a formal notion of privacy, introduced by
\citet{dwork2006calibrating}:

\begin{definition}[$(\epsilon, \delta)$-DP~\citep{dwork2006calibrating}]
	A randomized mechanism $\cM:\cS^n\rightarrow\cR$ satisfies $(\epsilon,\delta)$-differential privacy if for any two adjacent data
	sets $S,S'\in \cS^n$ differing by one element, and any output subset $O\subseteq \cR$, it holds that 
	\begin{align*}
	\PP[\cM(S)\in O]\leq e^\epsilon\cdot \PP[\cM(S')\in O]+\delta.
	\end{align*}
\end{definition}
 To achieve $(\epsilon, \delta)$-DP for a given function $q:\cS^n\rightarrow\cR$, we can use Gaussian mechanism \citep{dwork2014algorithmic} $\cM=q(S)+\ub$, where $\ub$ is a standard Gaussian random vector with variance that is proportional to the $\ell_2$-sensitivity of the function $q$, $\Delta(q)$, which is defined as follows.

\begin{definition}[$\ell_2$-sensitivity\citep{dwork2014algorithmic}]
	For two adjacent datasets $S,S'\in \cS^n$ differing by one element, the $\ell_2$-sensitivity $\Delta(q)$ of a function $q:\cS^n\rightarrow\cR$ is defined as 
\begin{align*}
    \Delta(q)=\sup_{S,S'}\|q(S)-q(S')\|_2.
\end{align*}
\end{definition}

\noindent\textbf{R\'enyi differential privacy.} Although the notion of $(\epsilon,\delta)$-DP is widely used in the output and objective perturbation methods, it suffers from the loose composition and privacy-amplification by subsampling results, which makes it unsuitable for the stochastic iterative learning algorithms. In this work, we will make use of the notion of R\'enyi Differential Privacy (RDP) \citep{mironov2017renyi} which is particularly useful when
the dataset is accessed by a sequence of randomized mechanisms \citep{wang2018subsampled}.

\begin{definition}[RDP~\citep{mironov2017renyi}]\label{def:rdp}
	For $\alpha>1,\rho>0$, a randomized mechanism $\cM:\cS^n\rightarrow\cR$ satisfies $(\alpha, \rho)$-R\'enyi differential privacy, i.e., $(\alpha, \rho)$-RDP, if for all adjacent datasets $S, S^\prime \in \cS^n$ differing by one element, we have  
	\begin{align*}
	 D_{\alpha}\big(\cM(S)||\cM(S^\prime)\big):=\log \EE\big[\big(\cM(S)/\cM(S^\prime)\big)^\alpha\big]/(\alpha-1)\leq \rho.
	\end{align*}
\end{definition}

According to Definition \ref{def:rdp}, RDP measures the ratio of probability distributions $\cM(S)$ and $\cM(S^\prime)$ by $\alpha$-order Renyi Divergence with $\alpha\in (1,\infty)$. As $\alpha\rightarrow\infty$, RDP reduces to $\epsilon$-DP. 

To further improve the privacy guarantee when using the Gaussian mechanisms to satisfy RDP, we establish the following privacy-amplification by subsampling result, which is derived based on the result in \citep{wang2018subsampled}.

\begin{lemma}
\label{lemma:GaussianM_RDP}
	Given a function $q:\cS^n\rightarrow\cR$, the Gaussian Mechanism $\cM=q(S)+\ub$, where $\ub\sim N(0,\sigma^2\Ib)$, satisfies $(\alpha,\alpha\Delta^2(q)/(2\sigma^2))$-RDP. In addition, if we apply the mechanism $\cM$ to a subset of samples using uniform sampling without replacement with sampling rate $\tau$, $\cM$ satisfies $(\alpha, 3.5\tau^2\Delta^2(q)\alpha/\sigma^2)$-RDP given $\sigma^{\prime2}=\sigma^2/\Delta^2(q)\geq 0.7$, $\alpha\leq 2\sigma^2\log(1/\big(\tau\alpha\big(1+\sigma^{\prime2})\big)\big)/3+1$.
\end{lemma}
\begin{remark}[\textit{Comparison with moment accountant}]\label{remark:sub_amp}
    Suppose $\Delta(q)=1$, Lemma \ref{lemma:GaussianM_RDP} suggests that to achieve $(\alpha, 3.5\tau^2\alpha/\sigma^2)$-RDP of the subsampled Gaussian mechanism, we require $\sigma^2\geq 0.7$. For the moment accountant based method \citep{abadi2016deep}, it can achieve the asymptotic privacy guarantee of $\big(\alpha, \tau^2\alpha/(1-
    \tau)\sigma^2+O(\tau^3\alpha^3/\sigma^3)\big)$-RDP when $\tau$ goes to zero and $\sigma^2\geq 1$, $\alpha\leq \sigma^2\log (1/(\tau\sigma))$. In contrast to moment accountant, our result has a closed-form bound on the privacy guarantee and a relaxed requirement of $\sigma^2$. 
\end{remark}
    It is worth noting that there exist some other works \citep{mironov2019r,zhu2019poission} also studying the privacy-amplification by subsampling results. However, they consider the Poisson subsampling approach, which is different from our uniform subsampling method.

Based on Lemma \ref{lemma:GaussianM_RDP}, we can establish a strong privacy guarantee of our method in terms of RDP, and then transfer it to $(\epsilon,\delta)$-DP using the following lemma.
\begin{lemma}[\citet{mironov2017renyi}]\label{lemma:RDP_to_DP}
	If a randomized mechanism $\cM: \cS^n\rightarrow\cR$ satisfies $(\alpha,\rho)$-RDP, then $\cM$ satisfies $(\rho+\log(1/\delta)/(\alpha-1),\delta)$-DP for all $\delta\in(0,1)$.
\end{lemma}




\section{Algorithm}\label{sec:alg}

Our proposed algorithm for differentially private nonconvex ERM, is illustrated in Algorithm \ref{alg:DPSRM}. 

\begin{algorithm}[!thp]
	\caption{Differentially Private Stochastic Recursive Momentum (DP-SRM)} \label{alg:DPSRM}
	\begin{algorithmic}[1]
		\INPUT $\btheta^0,T,G,L,\gamma,\beta,n_0$, privacy parameters $\epsilon,\delta$, accuracy for the first-order stationary point $\zeta$  
		\STATE Uniformly sample $b_0$ examples without replacement indexed by $\cB_0$
		\STATE Compute $\vb^0=\nabla F_{\cB_0}(\btheta^0)$, where $\nabla F_{\cB_{0}}(\btheta^{0})=\sum_{i\in\cB_{0}}\nabla f_i(\btheta^{0})/b_0$, draw $\ub^{0}\sim N(0,\sigma^2_0\Ib_d)$ with $\sigma^2_0=14TG^2\alpha/(\beta n^2\epsilon)$, $\alpha=\log(1/\delta)/\big((1-\beta)\epsilon\big)+1$
		\STATE Release the differentially private gradient estimator $\vb_p^{0}=\vb^{0}+\ub^{0}$
		\FOR{$t=0,1,2,\ldots, T-1$}
		\STATE $\btheta^{t+1}=\btheta^{t}-\eta_t\vb^t_p$, where $\eta_t=\min\big\{\zeta/(n_0L\|\vb^t_p\|_2),1/(2n_0L)\big\}$ 
		\STATE Uniformly sample $b$ examples without replacement indexed by $\cB_{t+1}$
		\STATE Compute $\vb^{t+1}=\nabla F_{\cB_{t+1}}(\btheta^{t+1})+(1-\gamma)\big(\vb^{t}_p-\nabla F_{\cB_{t+1}}(\btheta^{t})\big)$, draw $\ub^{t+1}\sim N(0,\sigma^2\Ib_d)$ with $\sigma^2=14T\big((1-\gamma)\zeta/n_0+\gamma G\big)^2\alpha/(\beta n^2\epsilon)$, $\alpha=\log(1/\delta)/\big((1-\beta)\epsilon\big)+1$
		\STATE Release the differentially private gradient estimator $\vb_p^{t+1}=\vb^{t+1}+\ub^{t+1}$
		\ENDFOR
		\OUTPUT $\tilde \btheta$ chosen uniformly at random from $\{\btheta^t\}_{t=0}^{T-1}$
	\end{algorithmic}
\end{algorithm}

The main idea is to construct the differentially private gradient estimator $\vb^t_p$ iteratively based on the information obtained from the previous updates. We initialize $\vb^0$ to be the mini-batch stochastic gradient $\nabla F_{\cB_0}(\btheta^0)$ and inject Gaussian noise, $\ub^0$,  with covariance matrix $\sigma^2_0 \Ib_d$ (lines 2, 3), to make it differentially private. Then, we recursively update $\vb^t$ (line 7) as $\vb^t=\nabla F_{\cB_{t}}(\btheta^{t})+(1-\gamma)\big(\vb^{t-1}_p-\nabla F_{\cB_{t}}(\btheta^{t-1})\big)$, where $\nabla F_{\cB_{t}}(\btheta^t)$, $\nabla F_{\cB_{t}}(\btheta^{t-1})$ are mini-batch stochastic gradients and $\vb^{t-1}_p$ is the private gradient estimator released at the last iteration. 
The momentum parameter, $\gamma$, is used to control the decay rate of the prior information, $\vb^{t-1}_p-\nabla F_{\cB_{t}}(\btheta^{t-1})$. This is called stochastic recursive momentum \citep{cutkosky2019momentum}, which can lead to fast convergence.
After updating $\vb^t$, we again inject Gaussian noise $\ub^t$ with covariance matrix $\sigma_2\Ib_d$ (line 8), to provide differential privacy. The variance $\sigma^2_0$, $\sigma^2$ of the Gaussian random vectors are determined by our RDP-based analysis. We choose an adaptive step size (line 5) to bound the sensitivity of the gradient estimator $\vb^t_p$, which is the key to establish the tight privacy and utility guarantees (Section \ref{sec:proof_outline}) of our algorithm. 


\section{Main Theoretical Results}\label{sec:results}

In this section, we establish formal privacy and utility guarantees for Algorithm~\ref{alg:DPSRM}. 

\begin{theorem}\label{thm:dp}
	Suppose that each component function $f_i$ is $G$-Lipschitz and has $L$-Lipschitz gradient. Given the total number of iterations $T$, the momentum parameter $\gamma$ and the accuracy for the first-order stationary point $\zeta$, for any $\delta>0$ and the privacy budget $\epsilon$, Algorithm \ref{alg:DPSRM} satisfies $(\epsilon,\delta)$-differential privacy with $\sigma_0^2=14TG^2\alpha/(\beta n^2\epsilon)$ and $\sigma^2=14T\big((1-\gamma)\zeta/n_0+\gamma G\big)^2\alpha/(\beta n^2\epsilon)$ if we have  $\alpha-1=\log(1/\delta)/\big((1-\beta)\epsilon\big)\leq 2\sigma^{\prime 2}\log\big(1/\big(\tau\alpha (1+\sigma^{\prime 2})\big)\big)/3$ with $\beta\in(0,1)$ and $\sigma^{\prime2}=\min\{b^2\sigma^2/\big(4((1-\gamma)\zeta/n_0+\gamma G)^2\big), b_0^2\sigma_0^2/(4G^2)\}\geq 0.7$, where $b_0$ and $b$ are batch sizes, and $\tau=\max\{b_0/n,b/n\}$.
\end{theorem}
\begin{remark}
According to Theorem \ref{thm:dp}, there exists a constraint on the parameter $\alpha$, which is due to the privacy-amplification by subsampling result in Lemma \ref{lemma:GaussianM_RDP}, and is similar to the constraint given by the moments accountant \citep{abadi2016deep} and other RDP-based analyses with subsampling approaches \citep{mironov2019r,zhu2019poission}. 
Furthermore, as we mentioned in Remark \ref{remark:sub_amp}, our result relaxes the requirement of the variance $\sigma^{\prime2}$  compared with the moments accountant based analysis.
\end{remark}

Following the previous work \citep{bassily2019private}, we can get rid of the constraints in Theorem \ref{thm:dp} by using a larger mini-batch size, as states in the following corollary.
\begin{corollary}\label{coro:dp}
	Given the total number of iterations $T$, the momentum parameter $\gamma$ and the accuracy for the first-order stationary point $\zeta$. Under the same conditions of Theorem \ref{thm:dp} on $f_i,\sigma_0^2,\sigma^2$, for any $\delta>0$ and the privacy budget $\epsilon$, Algorithm \ref{alg:DPSRM} satisfies $(\epsilon,\delta)$-differential privacy if we choose $b_0^2=b^2= n^2\epsilon/T$, $\beta=1/2$, and $T$  is larger than $O\big(\log^4(1/\delta)/\epsilon^3\big)$.
\end{corollary}

Theorem \ref{thm:dp} and Corollary \ref{coro:dp} require that each component function $f_i$ is $G$-Lipschitz and has $L$-Lipschitz gradient which will be used to derive the sensitivity of the underlying query function (i.e., the gradient estimator $\vb^t$ in Algorithm \ref{alg:DPSRM}) and thus determine the variance of the Gaussian noise. The $G$-Lipschitz condition has been widely assumed in the literature of differential privacy \citep{abadi2016deep,wang2017differentially,jayaraman2018distributed,bassily2019private}, and the $L$-Lipschitz gradient condition has also been made in several previous works~\citep{zhang2017efficient,feldman2020private}. In practice, we can use the clipping technique \citep{abadi2016deep} to ensure that at each iteration, $\|\nabla f_i(\btheta^t)\|_2\leq C_1$ and $\|\nabla f_i(\btheta^t)-\nabla f_i(\btheta^{t-1})\|_2\leq C_2$, where $C_1,C_2$ are some predefined constants. As a result, we can guarantee that the sensitivity of $\vb^t$ is bounded by $2\big((1-\gamma)C_2+\gamma C_1\big)/b$ (see \eqref{eq:sentivity}). Then, we can replace $G$ and $\zeta/n_0$ with $C_1$ and $C_2$ in Algorithm \ref{alg:DPSRM} to establish the same privacy guarantee.


The following theorem shows the utility guarantee and the gradient complexity, which is the total number of the stochastic gradients we need to estimate during the training process, of Algorithm \ref{alg:DPSRM}.
\begin{theorem}\label{thm:utility}
	Under the same conditions of Theorem \ref{thm:dp} on $f_i,\sigma^2,\sigma_0^2,\sigma^{\prime 2},\alpha$, if we choose the  
	number of iterations $T=C_1(n\epsilon LD_F)^{4/3}/\big( G^{8/3}(d\log(1/\delta)^{2/3}\big)$, where $D_F=F(\btheta^0)-F(\btheta^*)$ and $F(\btheta^*$) is a global minimum of $F$, the accuracy for the first-order stationary point  $\zeta=C_2\big(GLD_Fd\log(1/\delta)\big)^{1/3}/(n\epsilon)^{2/3}$, batch sizes $b_0=C_3G^3/(\zeta LD_F)$, $b=C_4G/(n_0\zeta)$, $n_0=LD_F/G^2$, the momentum parameter $\gamma^2=C_5\zeta^2/(n_0^2G^2)$ and $ n\epsilon\geq C_{6}\max\{G^{8}\log^{2}(1/\delta)/(LD_Fd)^{4},G^2(d\log(1/\delta))^{1/2}/(LD_F)\}$,  then the output $\tilde \btheta$ of Algorithm \ref{alg:DPSRM} and satisfies the following
	\begin{align*}
	\EE\|\nabla F(\tilde \btheta)\|_2 \leq C_7\bigg(\frac{\sqrt{GLD_Fd\log(1/\delta)}}{n\epsilon}\bigg)^{\frac{2}{3}},
	\end{align*}
	where $\{C_i\}_{i=1}^7$ are absolute constants, and the expectation is taken over all the randomness of the algorithm, i.e., the random Gaussian noise and the subsample gradient. Since $T= O\big((n\epsilon LD_F)^{4/3}/\big( G^{8/3}(d\log(1/\delta)^{2/3}\big)$, $b_0=b=O\big(G^{8/3}(n\epsilon)^{2/3}/(LD_F)^{4/3}(d\log(1/\delta))^{1/3}\big)$, the total gradient complexity of Algorithm \ref{alg:DPSRM} is $O\big((n\epsilon)^{2}/(d\log(1/\delta))\big)$.
	\end{theorem}
	
\begin{remark}[\textit{Comparison with existing methods}]
According to Theorem \ref{thm:utility}, our method can achieve the following utility guarantee
\begin{align*}
  O\bigg(\bigg(\frac{\sqrt{GLD_Fd\log(1/\delta)}}{n\epsilon}\bigg)^{\frac{2}{3}}\bigg).
\end{align*}
This result is better than the best known result for differentially private nonconvex optimization method \citep{wang2017differentially}. Furthermore, their method is based on gradient descent, which is computationally very expensive in large-scale machine learning problems.  Furthermore, the gradient complexity of our method is 
\begin{align*}
    O\bigg(\frac{(n\epsilon)^{2}}{d\log(1/\delta)}\bigg).
\end{align*}
This result is smaller than $O(n^2)$ gradient complexity provided by \citet{zhang2017efficient} and $O\big(n^2\epsilon/(d\log(1/\delta))^{1/2}\big)$ gradient complexity provided by \citet{wang2017differentially} when $d$ is large. 
\end{remark}

Theorem \ref{thm:utility} shows that our method only requires the computation of minibatch gradients with batch size at the order of $O\big((n\epsilon)^{2/3}/(d\log(1/\delta)^{1/3}\big)$ (ignoring the dependence on other parameters). Therefore, our method is more scalable than existing differentially private stochastic variance reduced algorithms, such as DP-SVRG \citep{wang2017differentially} designed for convex optimization, which often require the periodic computation of the checkpoint gradient with a giant batch size (full batch in DP-SVRG).

\section{Proof Outline of the Main Results}\label{sec:proof_outline}
In this section, we present the proof outline of the main results in Section \ref{sec:results}. 
Our proof involves new
techniques for the privacy and utility guarantees that are of general use for variance reduction-based algorithms. The detailed proof can be found in Section B in Appendix. 

\subsection{Privacy Guarantee}
According to Algorithm \ref{alg:DPSRM}, the mechanism at $t$-th iteration is $\cM_t$, which is a composition of $t$ Gaussian mechanisms: $\cG_{0},\ldots,\cG_{t}$, where $\cG_{0}=\nabla F_{\cB_0}(\btheta^0)+\ub^{0}$ and $\cG_t=\nabla F_{\cB_t}(\btheta^t)-(1-\gamma)\nabla F_{\cB_t}(\btheta^{t-1})+\ub^t$. Therefore, we want to show that $\cM_t$ is differentially private. For the given dataset $S$, we use $S^\prime$ to denote its neighboring dataset with one different example indexed by $i^\prime$ 

There are two main challenges in providing a tight privacy analysis. The first one is to deal with the subsampled mechanisms $\{\cG_i\}_{i=0}^{T-1}$. The second one is to control the sensitivity of $\cG_t$ when $t>0$. The first challenge can be addressed by our privacy-amplification by subsampling result (Lemma \ref{lemma:GaussianM_RDP}), which gives us a tight closed-form bound on the privacy guarantee. We can overcome the second challenge by using an adaptive stepsize, enabling us to use a small amount of random noise to achieve differential privacy.

According to Algorithm \ref{alg:DPSRM}, $\cG_t$ is the application of the following Gaussian mechanism $\tilde \cG_t$ to a subset of uniformly sampled examples, indexed by $\cB_t$
\begin{align*}
    \tilde \cG_t=
    \left\{
	\begin{array} {ll}
	\frac{1}{b}\sum_{i=1}^n\nabla f_i(\btheta^0)+\ub^0, & t=0\\
		\frac{1}{b}\sum_{i=1}^n\big(\nabla f_i(\btheta^t)-\phi\nabla f_i(\btheta^{t-1})\big)+\ub^t, & t>0,
	\end{array}
	\right.
\end{align*}
where $\phi=1-\gamma$. For $\tilde \qb_0=\sum_{i=1}^n\nabla f_i(\btheta^0)/b_0$ in $\tilde \cG_0$, the sensitivity $\Delta(\tilde \qb_0)$ is determined by
\begin{align*}
   \|\tilde\qb_0(S)-\tilde\qb_0(S^\prime)\|_2\leq\frac{1}{b}\|\nabla f_i(\btheta^{0})-\nabla f_{i^\prime}(\btheta^{0})\|_2\leq \frac{2G}{b_0}, 
\end{align*}
where the last inequality is due to $G$-Lipschitz of each component function. For $\tilde \qb_t=\sum_{i=1}^n\nabla f_i(\btheta^t)/b-(1-\gamma)\sum_{i=1}^n\nabla f_i(\btheta^{t-1})/b$ in $\tilde \cG_t$ when $t>0$, the sensitivity $\Delta(\tilde \qb_t)=\|\tilde \qb_t(S)-\tilde \qb_t(S^\prime)\|_2$ is determined by 
\begin{align}\label{eq:sentivity}
    &\frac{1-\gamma}{b}\|\nabla f_i(\btheta^{t})-\nabla f_{i}(\btheta^{t-1})+\nabla f_{i^\prime}(\btheta^{t})-\nabla f_{i^\prime}(\btheta^{t-1})\|_2
    +\frac{\gamma}{b}\|\nabla f_i(\btheta^{t})-\nabla f_{i^\prime}(\btheta^{t})\|_2.
\end{align}
Therefore, we have
\begin{align*}
  \|\qb_t(S)-\qb_t(S^\prime)\|_2&\leq \frac{2L(1-\gamma)}{b}\|\btheta^t-\btheta^{t-1}\|_2+\frac{2\gamma G}{b}\\
  &=\frac{2L(1-\gamma)}{b}\eta_{t-1}\|\vb^{t-1}_p\|_2+\frac{2\gamma G}{b}\\
  &\leq \frac{2(1-\gamma)\zeta}{n_0b}+\frac{2\gamma G}{b},
\end{align*}
where the first inequality is due to $L$-Lipschitz continuous gradient and $G$-Lipschitz of each component function. The last inequality comes from the adaptive stepsize $\eta_t=\min\big\{\zeta/(n_0L\|\vb^t_p\|_2),1/(2n_0L)\big\}$. Note that the proposed adaptive stepsize $\eta_t$ is the key to control the sensitivity of $\tilde \qb_t$. If we choose a fixed stepsize such as $\eta_t=1/(2L)$, the sensitivity of $\tilde \qb_t$ will be in the order of $O(G^2/b)$, which will lead to a much larger random noise to achieve differential privacy and thus deteriorate the utility of our method.

According to Lemma \ref{lemma:GaussianM_RDP}, if the noise $\ub^0$ and $\ub^t$ satisfy $\sigma^2_{0}=14T\alpha G^2/(\beta n^{2}\epsilon)$ and $\sigma^2=14T\alpha\big((1-\gamma)\zeta/n_0+\gamma G\big)^2/(\beta n^2\epsilon)$,
the Gaussian mechanism $\tilde \cG_{t}$ satisfies $\big(\alpha,\beta\epsilon n^2/\big(7b_0^2T\big)\big)$-RDP, and the privacy-amplification by subsampling result shows that $\cG_t$ satisfies $(\alpha,\beta\epsilon/T)$-RDP. Therefore, by the composition rule of RDP \cite{mironov2017renyi}, after $T^\prime$ iterations, Algorithm \ref{alg:DPSRM} satisfies $(\alpha,\beta T^\prime\epsilon/T)$-RDP. According to Lemma \ref{lemma:RDP_to_DP} and $\alpha=\log(1/\delta)/\big((1-\beta)\epsilon\big)+1$, we have that after $T^\prime$ iterations, Algorithm \ref{alg:DPSRM} satisfies $(T^\prime\epsilon/T,\delta)$-DP.

\subsection{Utility Guarantee}
According to the definition of $\tilde \btheta$, we have 
\begin{align*}
    \EE\|\nabla F(\tilde \btheta)\|_2&=\frac{1}{T}\sum_{t=0}^{T-1}\EE\|\nabla F(\btheta^t)\|_2\leq \frac{1}{T}\sum_{t=0}^{T-1}\EE\big\|\vb^t_p\big\|_2+\frac{1}{T}\sum_{t=0}^{T-1}\EE\big\|\nabla F(\btheta^t)-\vb^t_p\big\|_2,
\end{align*}
 where the expectation is taken over all the randomness of the algorithm. The key challenge in establishing a tight utility guarantee is to derive tight upper bounds for $\sum_{t=0}^{T-1}\EE\big\|\vb^t_p\big\|_2/T$ and $\sum_{t=0}^{T-1}\EE\big\|\nabla F(\btheta^t)-\vb^t_p\big\|_2/T$ when we have adaptive stepsize $\eta_t$ and the random noise $\ub^t$ in $\vb_p^t$.

First of all, by taking into account the adaptive stepsize $\eta_t$, we can upper bound the term $\sum_{t=0}^{T-1}\EE\big\|\vb^t_p\big\|_2/T$ as follows
\begin{align*}
 \frac{4n_0LD_F}{T\zeta}+\frac{1}{T\zeta}\sum_{t=0}^{T-1}\EE\big\|\nabla F(\btheta^t)-\vb^t_p\big\|_2^2+2\zeta,
\end{align*}
where $D_F=F(\btheta^0)- F(\btheta^*)$. Furthermore,  we can obtain the upper bound for $\sum_{t=0}^{T-1}\EE\big\|\vb^t_p-\nabla F(\btheta^t)\big\|_2^2/T$ as follows
\begin{align*}
     \frac{2(1-\gamma)^2\zeta^2}{n_0^2\gamma b}+\frac{2\gamma G^2}{b}+\frac{G^2}{T\gamma b_0}+\frac{Td\sigma^2+d\sigma_0^2}{T\gamma},
\end{align*}
 where the first term is determined by $\eta_t$, and the last term is determined by the random noise $\ub^t$ in $\vb_p^t$. The last term in this bound is dominated by $d\sigma^2/\gamma$, which validates the necessity of using the adaptive stepsize to control the sensitivity of $\vb^t$ and thus enable a small $\sigma^2$.

Finally, combining these two new bounds and plugging the value of parameters in Theorem \ref{thm:utility}, we can obtain that
\begin{align*}
    \EE\|\nabla F(\tilde \btheta)\|_2\leq C_1\zeta+C_2\frac{\sqrt{GLD_Fd\log(1/\delta)}}{n\epsilon\sqrt{\zeta}}.
\end{align*}
By solving the smallest $\zeta$, we can obtain $\zeta=(GLD_Fd\log(1/\delta))^{1/3}/(n\epsilon C_1/C_2)^{2/3}$. Thus we have $\EE\|\nabla F(\tilde \btheta)\|_2\leq C_3\zeta$, where $C_1,C_2,C_3$ are some constants.

\section{Experiments}\label{sec:exp}
 This section presents results from experiments that evaluate our method's performance on different nonconvex ERM problems and different datasets. All experiments are implemented in
Pytorch platform version 1.2.0 within Python 3.7.6. on a local machine which comes with Intel Xeon 4214 CPUs and NVIDIA GeForce RTX 2080Ti GPU (11G GPU RAM).  

\subsection{Nonconvex Logistic Regression}
We first consider the binary logistic regression problem with a nonconvex regularizer \citep{reddi2016fast}
\begin{align*}
    \min_{\btheta\in\RR^d}&\frac{1}{n}\sum_{i=1}^ny_i\log\phi(\xb_i^\top\btheta)+(1-y_i)\log\big[1-\phi(\xb_i^\top\btheta)\big]+\lambda\sum_{i=1}^d\theta_j^2/(1+\theta_j^2),
\end{align*}
where $\phi(x)=1/\big(1+\exp(-x)\big)$ is the sigmoid function, $\theta_j$ is the $j$-th coordinate of $\btheta$, and $\lambda>0$ is the regularization parameter. We set $\lambda=0.001$ in this experiment. 
Here, we consider two commonly-used binary classification benchmark datasets: \textit{a9a} dataset, which contains 32561 training examples, 16281 test examples, 123 features, and \textit{ijcnn1} dataset with 49990 training examples, 91701 test examples, 22 features.

\shortsection{Baseline methods} We compare our method (DP-SRM) with random round private stochastic gradient descent (RRPSGD) proposed by \citet{zhang2017efficient} , differentially private gradient descent (DP-GD) proposed by \citet{wang2017differentially}, and differentially private adaptive gradient descent (DP-AGD) proposed by \citet{lee2018concentrated}. 

\shortsection{Gradient clipping and privacy tracking}
We use the gradient clipping technique of \citet{abadi2016deep} to ensure that at $t$-th iteration of Algorithm \ref{alg:DPSRM}, $\|\nabla f_i(\theta^t)\|_2$ and $\|\nabla f_i(\theta^t)-\nabla f_i(\theta^{t-1})\|_2$ are upper bounded by some predefined values $C_1$ and $C_2$, respectively. This will ensure that the sensitivity of the gradient estimator $\vb^t$ is upper bounded by $2\big((1-\gamma)C_1+\gamma C_2\big)$ (see \eqref{eq:sentivity}), and gives us the desired privacy protection. At each iteration, we add the Gaussian noise with variance $\sigma^2$, and keep track of the RDP according to Lemma \ref{lemma:GaussianM_RDP} and transfer it to $(\epsilon,\delta)$-DP according to Lemma \ref{lemma:RDP_to_DP}.

\shortsection{Parameters} For all the algorithms, the step size is tuned around the theoretical values to give the fastest convergence using grid search. For our method, we tune the batch size $b$ by searching the grid $\{50,100,200\}$. We set $C_1=1,C_2=0.01$ and $\gamma=C_2$.  We choose $\epsilon\in\{0.2,0.5\}$ and $\delta=10^{-5}$.

\begin{figure*}[t!]%
	\centering
	\subfigure[$\epsilon=0.2$]{
		\label{fig1:subfig:1.a} 
		\includegraphics[width=0.23\textwidth]{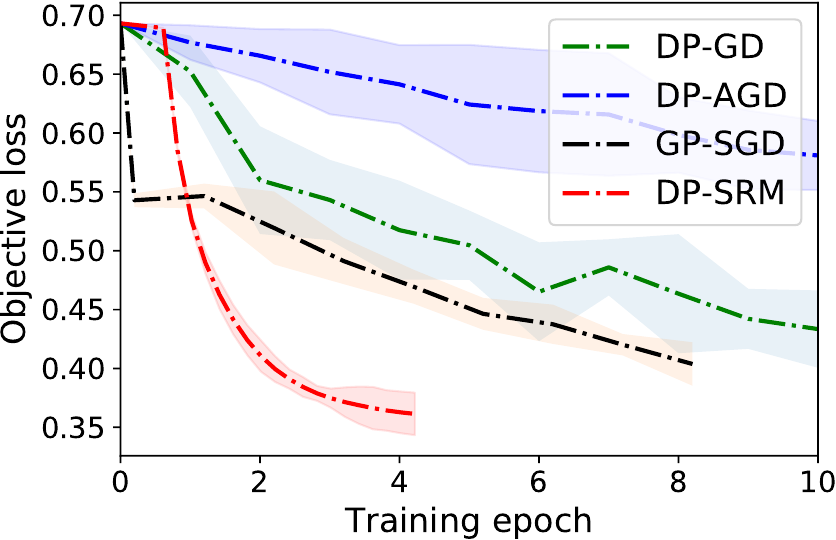}}
	\subfigure[$\epsilon=0.5$]{
		\label{fig1:subfig:1.b} 
		\includegraphics[width=0.23\textwidth]{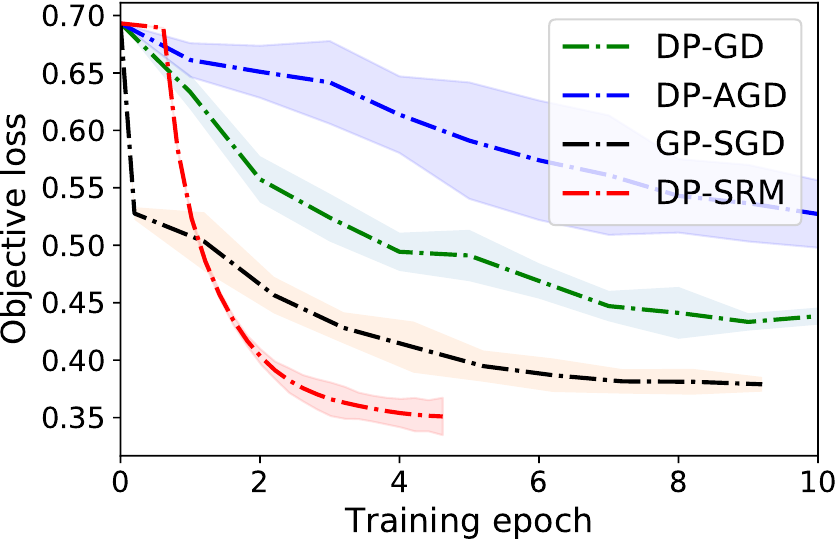}}
			\subfigure[$\epsilon=0.2$]{
		\label{fig1:subfig:1.c} 
		\includegraphics[width=0.23\textwidth]{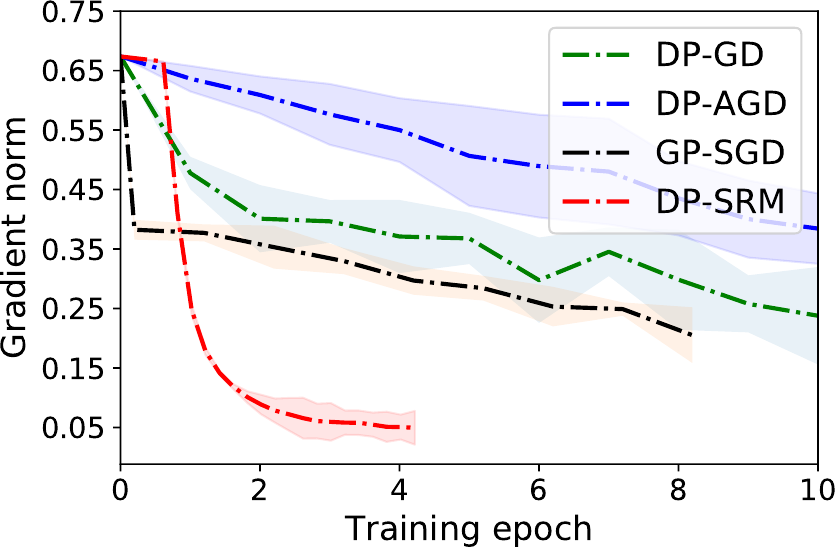}}
	\subfigure[$\epsilon=0.5$]{
		\label{fig1:subfig:1.d} 
		\includegraphics[width=0.23\textwidth]{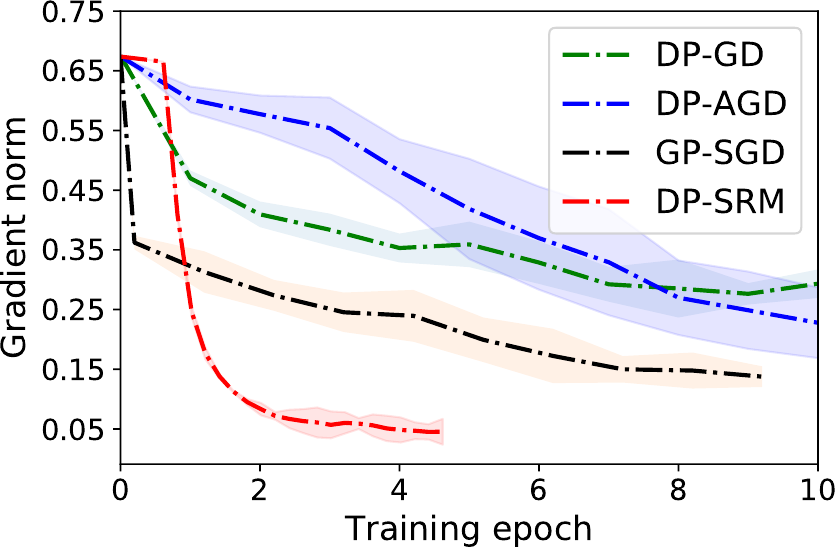}}

	\caption{Results for nonconvex logistic regression on \textit{a9a} dataset. (a), (b) illustrate the objective loss versus the number of epochs. (c), (d) present the gradient norm versus the number of epochs. } \label{figure:a9a}
\end{figure*}

\begin{figure*}[!th]%
	\centering
	\subfigure[$\epsilon=0.2$]{
		\label{fig3:subfig:1.a} 
		\includegraphics[width=0.23\textwidth]{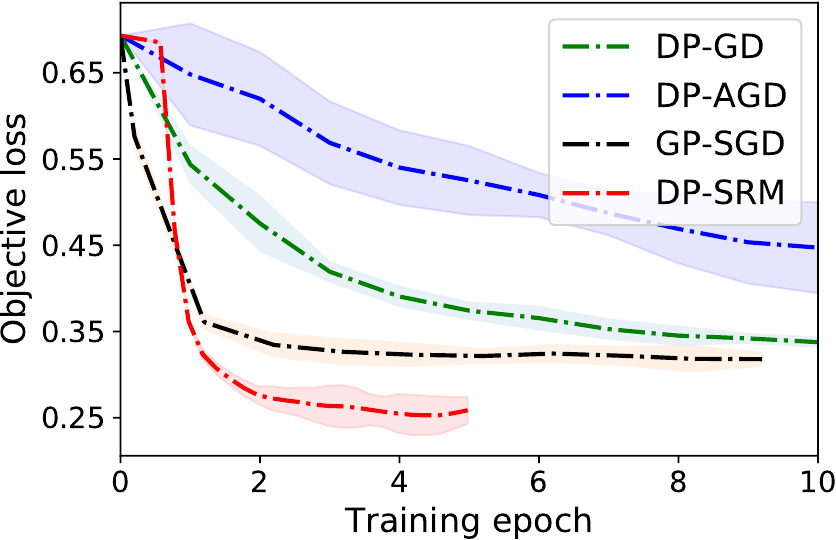}}
	\subfigure[$\epsilon=0.5$]{
		\label{fig3:subfig:1.b} 
		\includegraphics[width=0.23\textwidth]{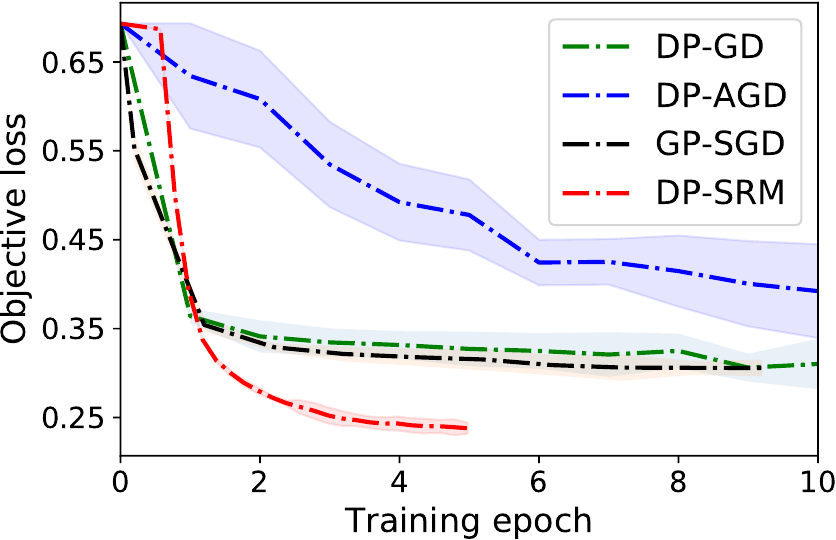}}
			\subfigure[$\epsilon=0.2$]{
		\label{fig3:subfig:1.c} 
		\includegraphics[width=0.23\textwidth]{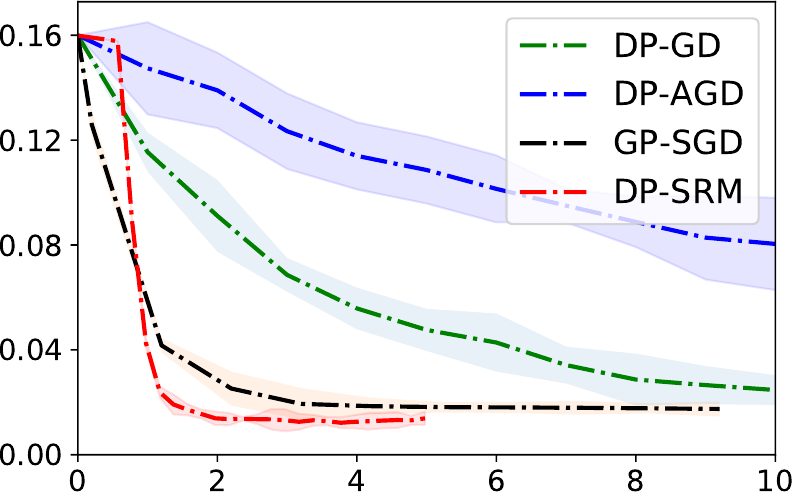}}
	\subfigure[$\epsilon=0.5$]{
		\label{fig3:subfig:1.d} 
		\includegraphics[width=0.23\textwidth]{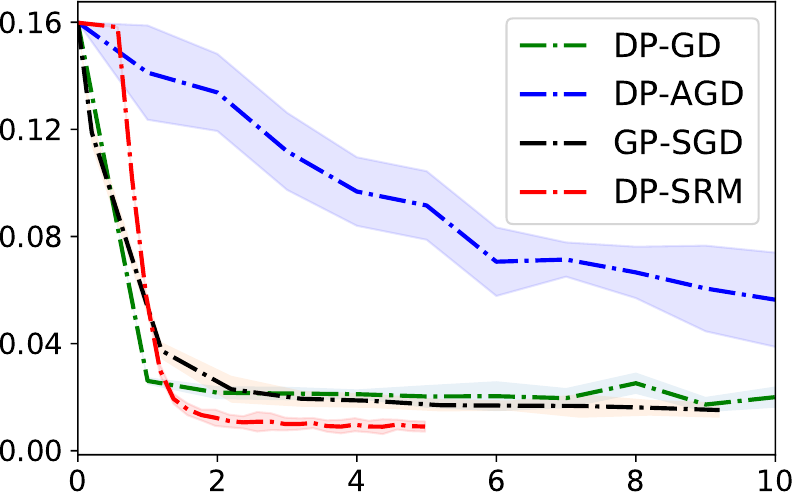}}

	\caption{Results for nonconvex logistic regression on \textit{ijcnn1} dataset. (a), (b) show the objective loss versus the number of epochs. (c), (d) illustrate the gradient norm versus the number of epochs. } \label{figure:ijcnn1}
\end{figure*}

\shortsection{Results}
Due to the randomized nature of all the algorithms, the experimental results are obtained by averaging the results over 30 runs. Figures~\ref{figure:a9a} and  \ref{figure:ijcnn1} show the objective function value and the gradient norm of different algorithms for privacy budgets $\epsilon\in\{0.2,0.5\}$ on \textit{a9a} and \textit{ijcnn1} datasets, respectively. We also report the 95\% confidence interval of these results. We can see from the plots that Our DP-SRM algorithm outperforms the other three baseline algorithms in terms of the objective loss, gradient norm, and convergence rate by a large margin. Tables~\ref{table:a9a} and \ref{table:ijcnn1} summarize the test error of different algorithms as well as the CPU time (in seconds) of the training process. The results also corroborate the advantages of our method in terms of accuracy and efficiency.

\begin{table*}[!t]
 \small
	\caption{Comparison of different algorithms on \textit{a9a} dataset when $\epsilon\in\{0.2,0.5\}$ and $\delta=10^{-5}$. We use the STORM algorithm \citep{cutkosky2019momentum} as the non-private baseline.}
	\label{table:a9a}
	\centering
	\begin{tabular}{l|c|c|c|c|c|c}
		\toprule 
		Privacy&  Non-private& \multirow{2}{*}{Method}&\multirow{2}{*}{Test Error} &Data& \multirow{2}{*}{CPU time (s)}  & \multirow{2}{*}{Gradient Norm}\\Budget& Baseline && & Passes &&\\
		\midrule
		\multirow{4}{*}{$\epsilon=0.2$}        & \multirow{4}{*}{0.3346} & DP-GD &    0.4155 (0.0107) &20  &  1.245   &    0.0953 (0.0212)  \\&&DP-AGD&0.3713 (0.0043) &360&96.21&0.0437 (0.0020)\\&&RRPSGD&0.4019 (0.0033) &8&39.61&0.2175 (0.0116)\\&(0.007)&\textbf{DP-SRM}&\textbf{0.3579 (0.0009)}&\textbf{4}&\textbf{0.6007}&\textbf{0.0528 (0.0042)}\\
		\midrule
		\multirow{4}{*}{$\epsilon=0.5$ }     & \multirow{4}{*}{0.3346}  &DP-GD & 0.3859 (0.0057)  &20 &   1.261  &     0.0866 (0.0129) \\&&DP-AGD&0.3627 (0.0038) &365&95.45 &0.0402 (0.0022)\\&&RRPSGD&0.3861 (0.0028)&10&52.32 &0.1454 (0.0126)\\&(0.007)&\textbf{DP-SRM}&\textbf{0.3506 (0.0011)}&\textbf{5}&\textbf{0.7383 }&\textbf{0.0502 (0.0061)}\\
		\hline
	\end{tabular}
\end{table*}

\begin{table*}[!th]
 \small
	\caption{Comparison of different algorithms on \textit{ijcnn1} dataset under different privacy budgets $\epsilon\in\{0.2,0.5\}$ and $\delta=10^{-5}$. Note that the non-private baseline denotes the test error of the non-private STORM algorithm \citep{cutkosky2019momentum}. }
	\label{table:ijcnn1}
	\centering
	\begin{tabular}{l|c|c|c|c|c|c}
		\toprule 
		Privacy&  Non-private& \multirow{2}{*}{Method}&\multirow{2}{*}{Test Error} &Data& \multirow{2}{*}{CPU time}  & \multirow{2}{*}{Gradient Norm}\\Budget& Baseline && & Passes &&\\
		\midrule
		\multirow{4}{*}{$\epsilon=0.2$}        & \multirow{4}{*}{0.2096} & DP-GD &    0.3160 (0.0120) &20  &  0.5180   &    0.0184 (0.0024)  \\&&DP-AGD&0.2645 (0.0044) &346&90.05&0.0133 (0.0018)\\&&RRPSGD&0.3110 (0.0106) &8&47.64&0.0175 (0.0023)\\&(0.002)&\textbf{DP-SRM}&\textbf{0.2503 (0.0090)}&\textbf{4}&\textbf{0.4748 }&\textbf{0.0117 (0.0008)}\\
		\midrule
		\multirow{4}{*}{$\epsilon=0.5$ }     & \multirow{4}{*}{0.2096}  &DP-GD & 0.2717 (0.0081)  &20 &  0.4990  &     0.0171 (0.0024) \\&&DP-AGD&0.2416 (0.0029) &365&94.28 &0.0397 (0.0025)\\&&RRPSGD&0.3033 (0.0110)&10&59.06 &0.0160 (0.0018)\\&(0.002)&\textbf{DP-SRM}&\textbf{0.2341 (0.0042)}&\textbf{5}&\textbf{0.4368 }&\textbf{0.0082 (0.0005)}\\
		\hline
	\end{tabular}
\end{table*}

\begin{figure*}[!t]%
	\centering
	\subfigure[$\epsilon=3.0$]{
		\label{fig2:subfig:1.a} 
		\includegraphics[width=0.23\textwidth]{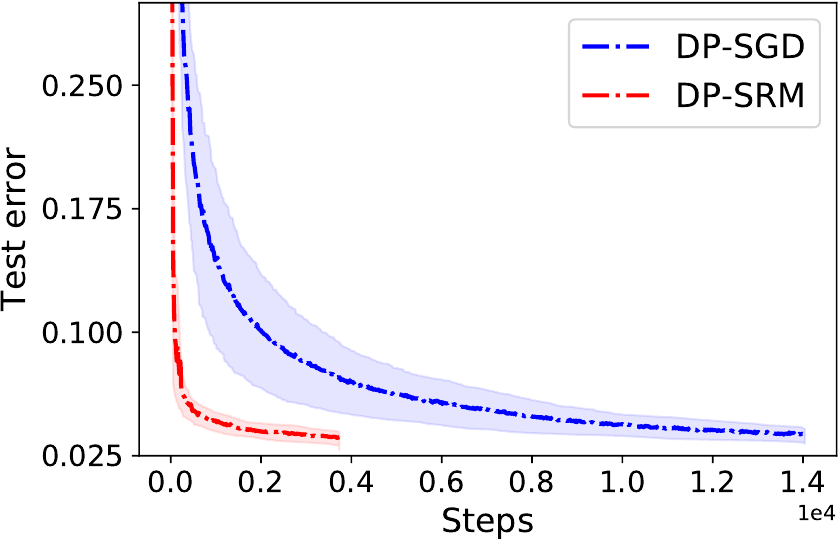}}
	\subfigure[$\epsilon=3.0$]{
		\label{fig2:subfig:1.b} 
		\includegraphics[width=0.23\textwidth]{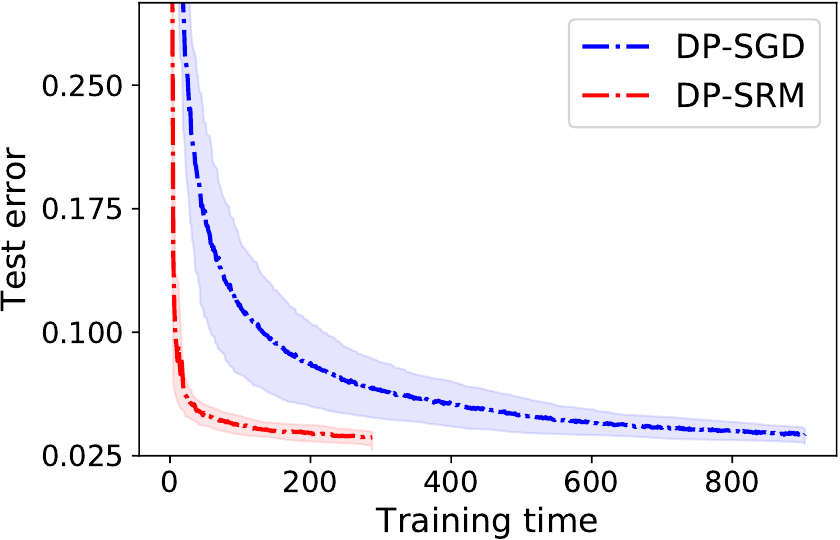}}
	\subfigure[$\epsilon=1.2$]{
		\label{fig2:subfig:1.c} 
		\includegraphics[width=0.23\textwidth]{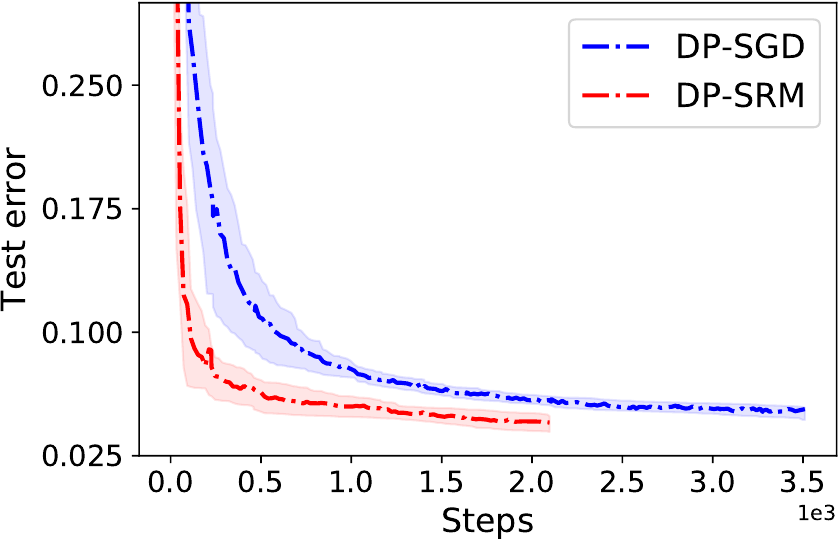}}
			\subfigure[$\epsilon=1.2$]{
		\label{fig2:subfig:1.d} 
		\includegraphics[width=0.23\textwidth]{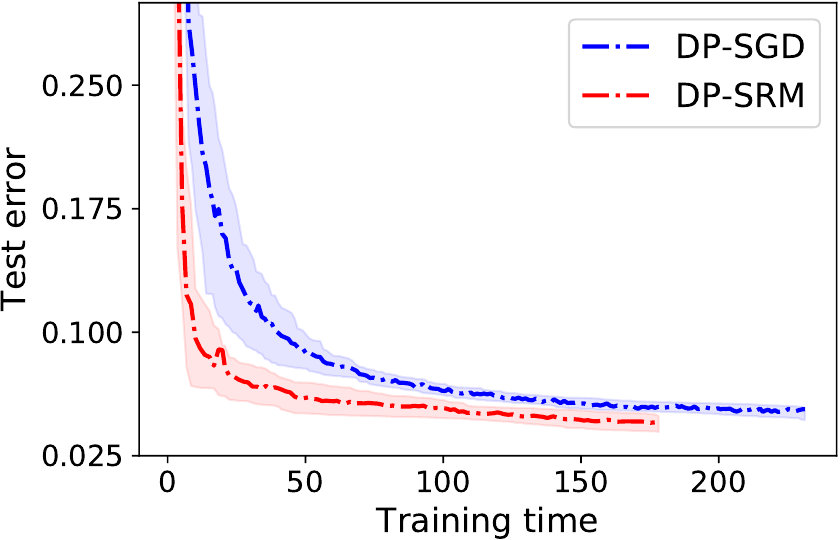}}
	\caption{Results on MNIST dataset. (a), (b) depict the test error under the privacy budget $\epsilon=3.0$. (c), (d) illustrate the test error under the privacy budget $\epsilon=1.2$.} \label{fig:mnist}
\end{figure*}
 \begin{figure*}[!thb]%
	\centering
	\subfigure[$\epsilon=2.0$]{
		\label{fig6:subfig:1.a} 
		\includegraphics[width=0.23\textwidth]{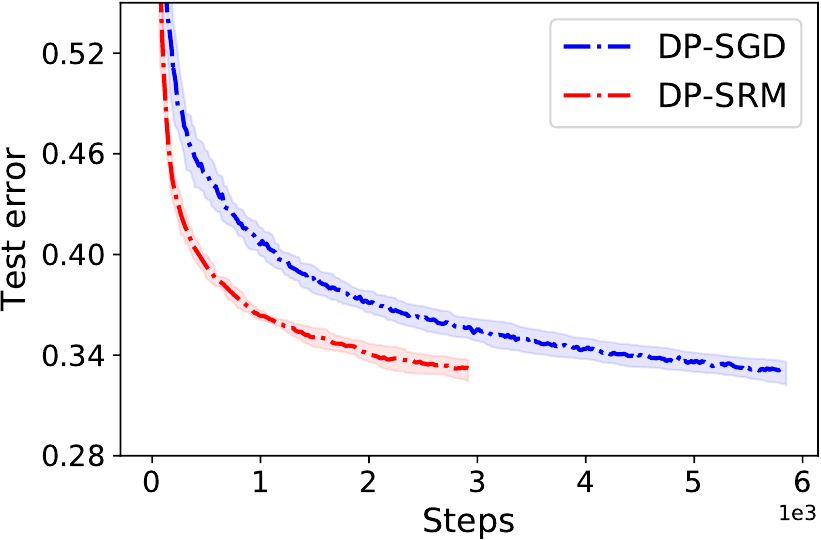}}
	\subfigure[$\epsilon=2.0$]{
		\label{fig6:subfig:1.b} 
		\includegraphics[width=0.23\textwidth]{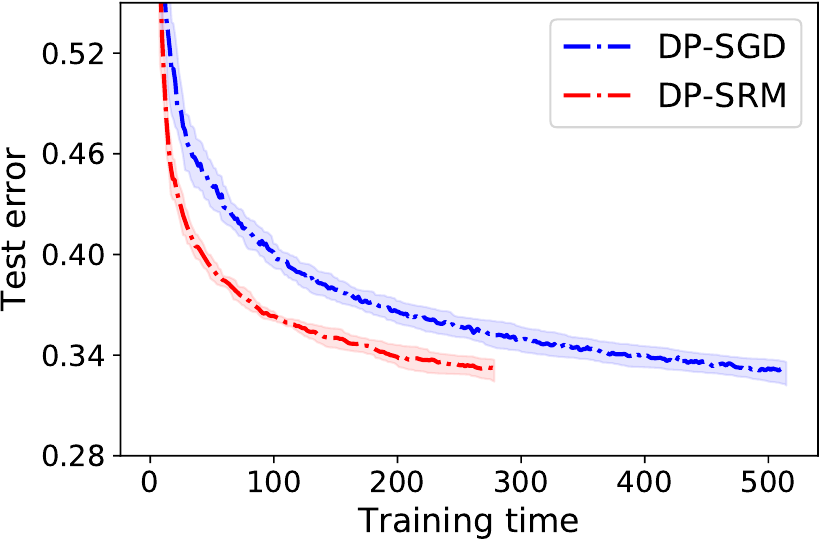}}
	\subfigure[$\epsilon=4.0$]{
		\label{fig6:subfig:1.c} 
		\includegraphics[width=0.23\textwidth]{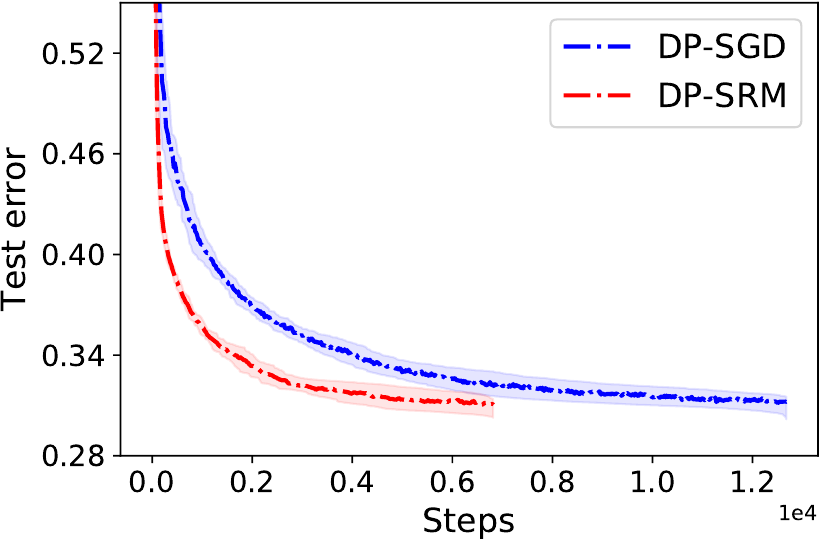}}
			\subfigure[$\epsilon=4.0$]{
		\label{fig6:subfig:1.d} 
		\includegraphics[width=0.23\textwidth]{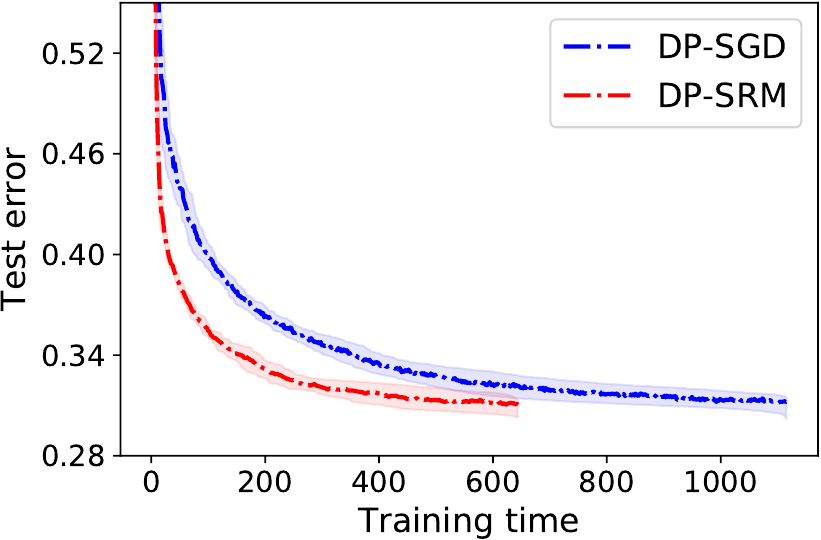}}
	\caption{Results for CNN6 on CIFAR-10 dataset. (a), (b) depict the test error under the privacy budget $\epsilon=2.0$. (c), (d) illustrate the test error under the privacy budget $\epsilon=4.0$} \label{fig:cifar10}
\end{figure*}

\subsection{Convolutional Neural Networks}
We compare our algorithm with the differentially private stochastic gradient descent (DP-SGD) algorithm proposed by \citet{abadi2016deep} on training convolutional neural networks for image classification on both MNIST \citep{lecun1998gradient} and CIFAR-10 \citep{krizhevsky2009learning} datasets. 

\shortsection{Architecture for MNIST} 
For MNIST dataset, we consider a 4 layer CNN \footnote{\url{https://github.com/facebookresearch/pytorch-dp}.}, which can achieve 99\% classification accuracy on the test dataset after training with SGD.

\shortsection{Parameters for MNIST} 
 We choose privacy budgets $\epsilon\in\{1.2,3.0,7.0\}$, and set $\delta=10^{-5}$. To ensure the privacy guarantee (see \eqref{eq:sentivity}), we set the clipping parameter $C_1=1.5$ for the term $\|\nabla f_i(\theta^t)\|_2$. For the term $\|\nabla f_i(\theta^t)-\nabla f_i(\theta^{t-1})\|_2$, we choose the clipping parameter $C_2$ from the grid $\{0.01,0.1,0.3,0.5,0.7,0.9,0.99\}$. 
 For both DP-SGD and DP-SRM, we tune the batch size $b$ by searching the grid $\{256, 512, 1024\}$ and the step size by $\{0.01,0.05,0.1,0.25,0.5\}$. For DP-SRM, we tune the batch size $b_0$ by $\{b,2b,4b\}$. In addition, we set the momentum parameter $\gamma=C_2$. 
 
\shortsection{Results for MNIST}
 Figures \ref{fig:mnist} illustrates the average test error and the corresponding 95\% confidence interval of different methods versus the number of iterations as well as the training time (in seconds) under the privacy budgets $\epsilon=1.2$ and $\epsilon=3.0$ over 30 trials. We see similar results under the privacy budget $\epsilon=7.0$, and thus defer them in Section A in Appendix. The CNN trained by the non-private SGD can achieve $1\%$ test error after 20 epochs. Figure \ref{fig2:subfig:1.a} and  Figure \ref{fig2:subfig:1.c}  show that our proposed method can achieve $3.62\%$ and $4.49\%$ test errors when $\epsilon=3.0$ and $\epsilon=1.2$, which are better than DP-SGD with $3.81\%$ and $5.33\%$ test errors. Besides,  our method converges faster than DP-SGD. Figure \ref{fig2:subfig:1.a} and Figure \ref{fig2:subfig:1.b} demonstrate that compared with DP-SGD, our method only takes $0.3\times$ iterations and $0.4\times$ training time to achieve comparable performances under the privacy budget $\epsilon=3.0$. 

\shortsection{Architecture for CIFAR-10} 
We consider two convolutional neural networks for CIFAR-10. The first one is a five layer CNN with two convolutional layers and three fully connected layers, and we call it CNN5 \footnote{\url{https://pytorch.org/tutorials/beginner/blitz/cifar10_tutorial.html}.}. For CNN5, we train it from the scratch using our DP-SRM method and the DP-SGD method \citep{abadi2016deep} and compare their performances in terms of the model accuracy, iteration numbers and the training time. For the second one, we consider a similar architecture as in \citet{abadi2016deep}, which has three convolutional layers with 32, 64, 128 filters in each convolution layer and three fully connected layers, and we denote it by CNN6. For CNN6, we follow the same experiment setting as in \citet{abadi2016deep}: we use CIFAR-100 dataset as a public dataset, and first train a network with the same architecture on this dataset as the pretrained model. Then, we initialize the convolutional layers of CNN6 using the cnvolutional layers of the pretrained model, and only train the fully connected layers of CNN6 on CIFAR-10 dataset using different private methods.

\shortsection{Parameters for CNN6} We choose three different privacy budgets $\epsilon\in\{2.0,4.0,8.0\}$ and $\delta=10^{-5}$. We set the clipping parameter $C_1=2$ for the term $\|\nabla f_i(\theta^t)\|_2$. For the term $\|\nabla f_i(\theta^t)-\nabla f_i(\theta^{t-1})\|_2$, we choose the clipping parameter $C_2$ by searching the grid $\{0.01,0.05,0.1,0.3,0.5,0.7,0.9,0.95,0.99\}$. 
For DP-SGD, we tune the step size by searching the grid $\{0.01,0.02,0.05,0.1,0.15,0.2\}$ and the batch size by $\{64, 128, 256\}$. For DP-SRM, we tune the batch size $b$ by searching the grid $\{64, 128, 256\}$, step size by $\{0.01,0.02,0.05,0.1,0.15,0.2\}$, and $b_0$ by $\{b,2b,4b\}$. In addition, we set the momentum parameter $\gamma=C_2$.

\shortsection{Results for CNN6} Figure \ref{fig:cifar10} presents the average test error and the corresponding 95\% confidence interval of different methods versus the number of iterations as well as the training time (in seconds) over 30 trials. The CNN6 trained by the non-private SGD will have $18.5\%$ test error after 150  epochs. The results show that our proposed method can achieve $33.2\%$ and $31.0\%$ test errors given $\epsilon=2.0$ and $\epsilon=4.0$, which are comparable to the results of DP-SGD with $33.2\%$ and $31.2\%$ under the same privacy budgets. However, we can see from the plots that our method can significantly reduce the iteration numbers and the training time. For example, when $\epsilon=4.0$, DP-SGD takes $1.3\times 10^{4}$ iterations and 1115 seconds to achieve $31.2\%$ test error. In sharp contrast, our method only takes $6.8\times 10^{3}$ iterations and 643 seconds to achieve $31.0\%$ test error. We can observe similar results for CNN5, which are presented in Section A in Appendix.

\section{Conclusions}\label{sec:8}
We propose an efficient differentially private algorithm for nonconvex ERM. We prove both privacy and utility guarantees for our method. Both theoretical analyses and experiments demonstrate the advantage of our algorithms compared with the state-of-the-art. It would be very interesting to study our method's performances in super large or even industrial level neural networks. It would also be very interesting to study the optimization lower bound for the differentially private nonconvex stochastic optimization problem.

\section*{Acknowledgements}

This work was supported by grants from the National Science Foundation (\#1717950 and \#1915813) and research awards from Baidu and Intel, and cloud computing grants from Amazon. The views and conclusions contained in this paper are those of the authors and should not be interpreted as representing any funding agencies.

\urlstyle{sf}
\bibliographystyle{ims}
\bibliography{arxiv_bib}

\appendix
\section{Additional Experiments}\label{sec:add_experiments}
\begin{figure*}[!thb]%
	\centering
	\subfigure[$\epsilon=7.0$]{
		\label{fig4:subfig:1.a} 
		\includegraphics[width=0.23\textwidth]{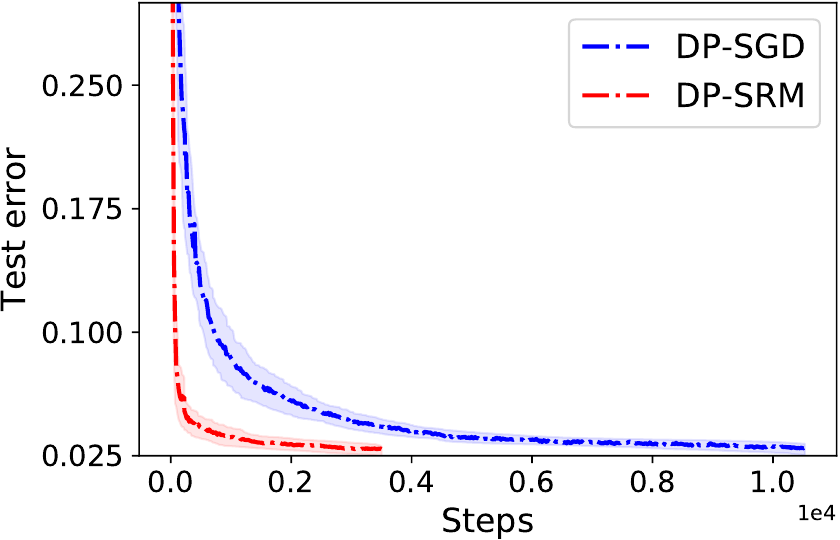}}
	\subfigure[$\epsilon=7.0$]{
		\label{fig4:subfig:1.b} 
		\includegraphics[width=0.23\textwidth]{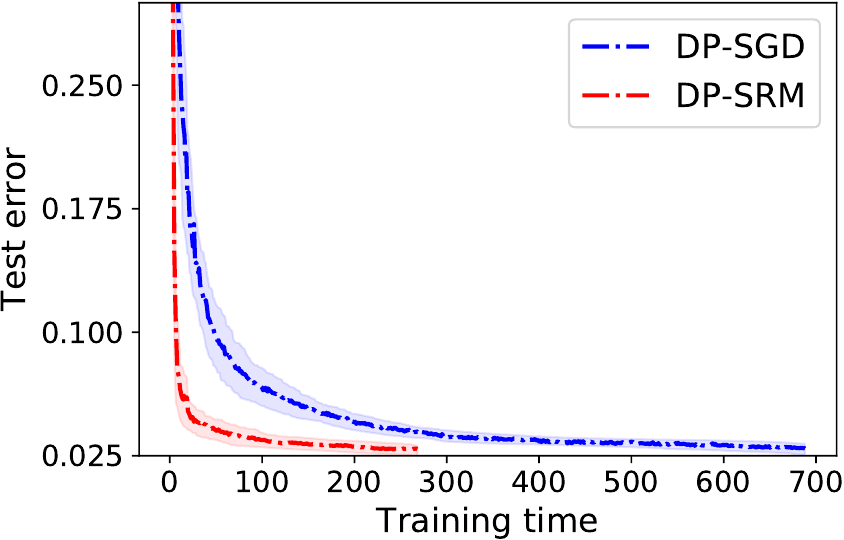}}
	\subfigure[$\epsilon=8.0$]{
		\label{fig4:subfig:1.c} 
		\includegraphics[width=0.23\textwidth]{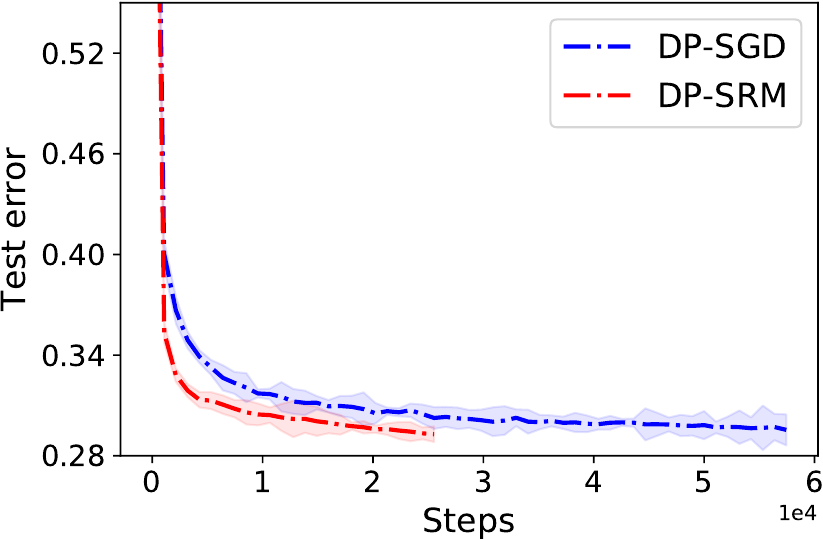}}
	\subfigure[$\epsilon=8.0$]{
		\label{fig4:subfig:1.d} 
		\includegraphics[width=0.23\textwidth]{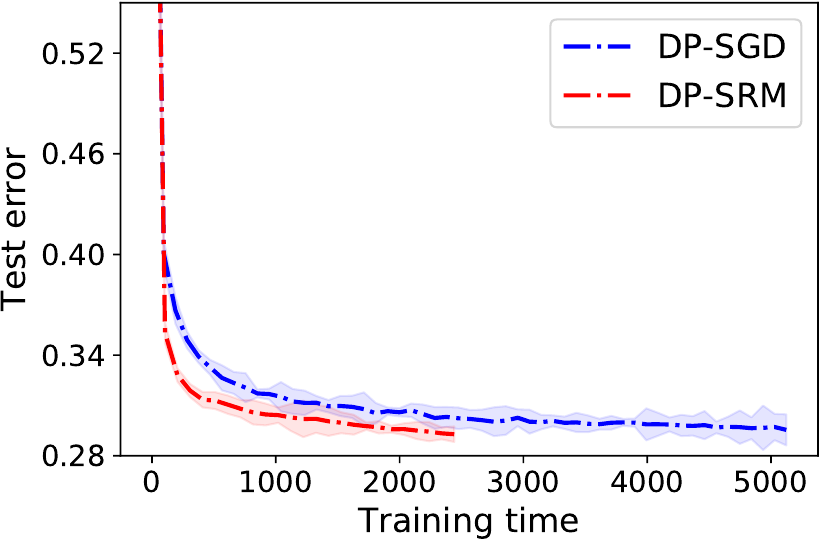}}
	\subfigure[$\epsilon=6.0$]{
		\label{fig4:subfig:1.e} 
		\includegraphics[width=0.23\textwidth]{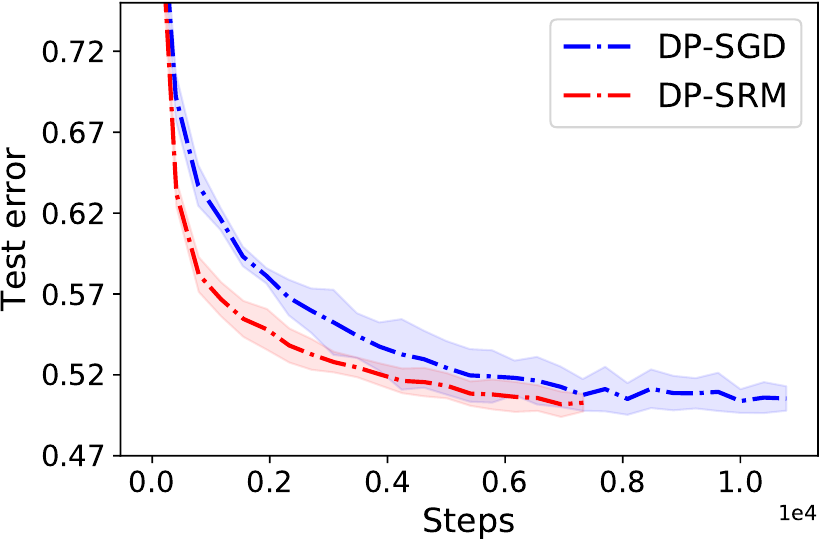}}
	\subfigure[$\epsilon=6.0$]{
		\label{fig4:subfig:1.f} 
		\includegraphics[width=0.23\textwidth]{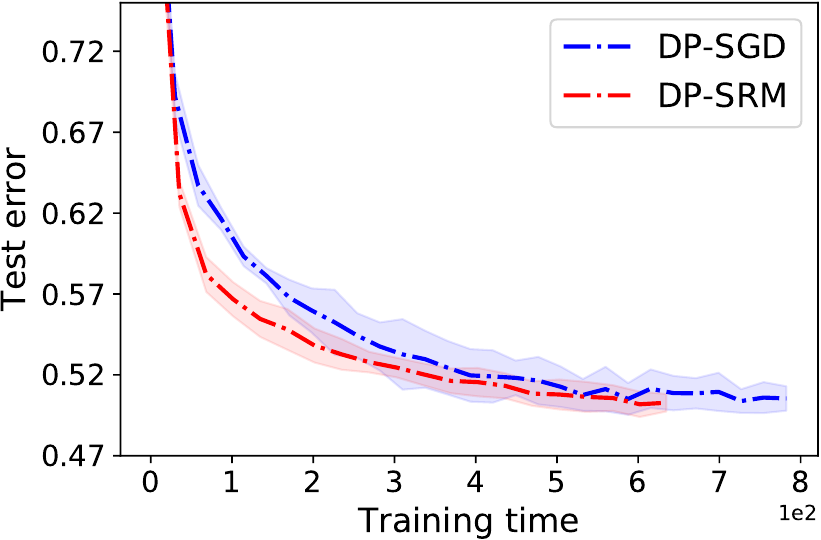}}
			\subfigure[$\epsilon=8.0$]{
		\label{fig6:subfig:1.h} 
		\includegraphics[width=0.23\textwidth]{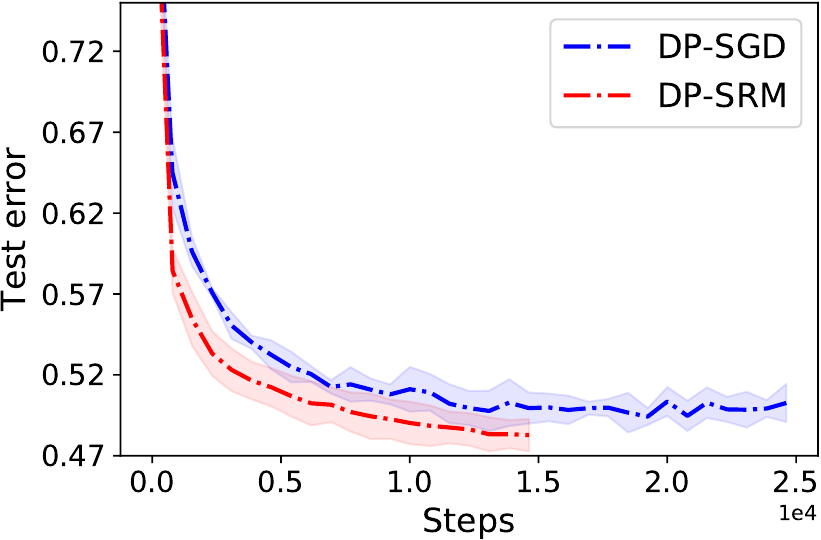}}
	\subfigure[$\epsilon=8.0$]{
		\label{fig6:subfig:1.h} 
		\includegraphics[width=0.23\textwidth]{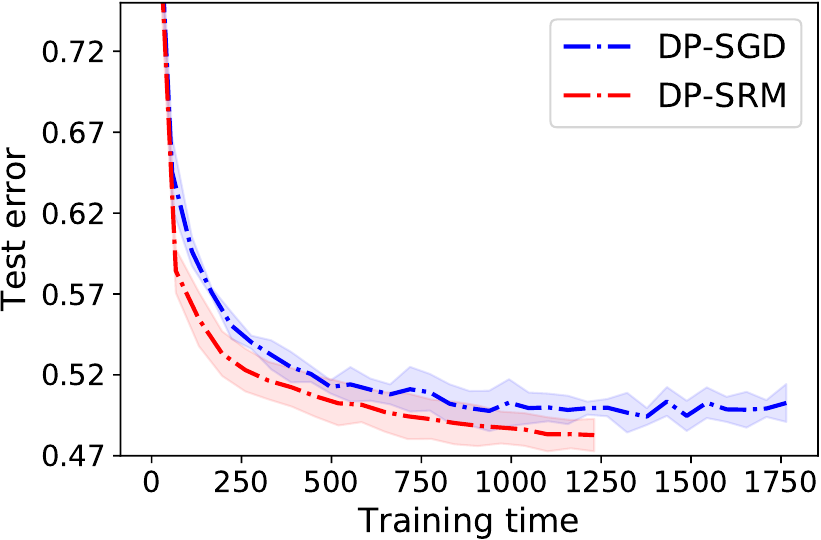}}
	\subfigure[$\epsilon=10.0$]{
		\label{fig6:subfig:1.i} 
		\includegraphics[width=0.23\textwidth]{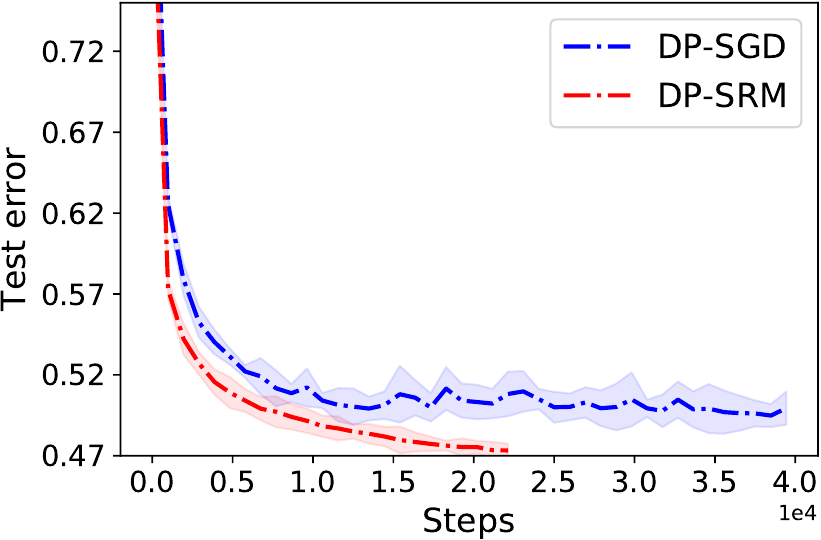}}
			\subfigure[$\epsilon=10.0$]{
		\label{fig6:subfig:1.j} 
		\includegraphics[width=0.23\textwidth]{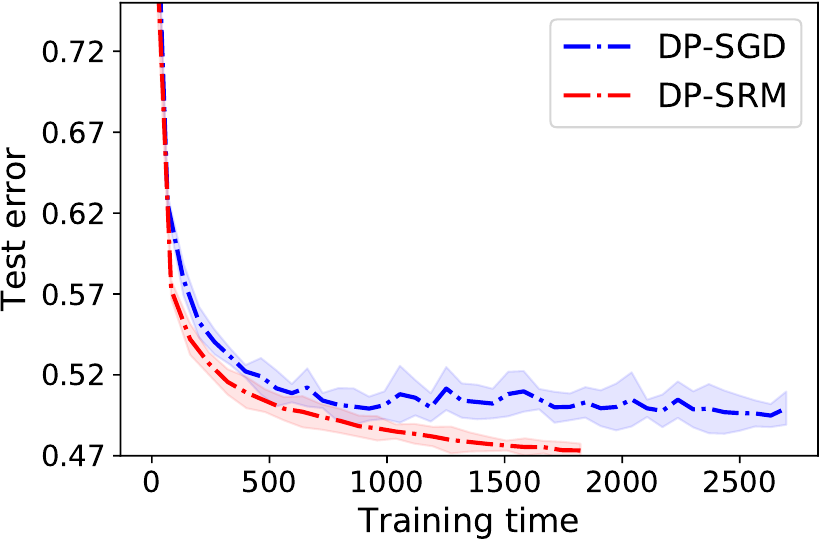}}
	\caption{Results for CNN on MNIST and CIFAR-10 datasets. (a), (b) illustrate the results on MNIST dataset. (c), (d) demonstrate the results for CNN6 on CIFAR-10 dataset. (e)-(j) show the results for CNN5 on CIFAR-10 dataset.} \label{fig:mnist_add}
\end{figure*}

In this section, we present additional experiment results on training convolutional neural networks. Figures \ref{fig:mnist_add} shows the average test error (over 30 trials) and the corresponding 95\% confidence interval of different methods versus the number of iterations as well as the training time under different privacy budgets on MNIST and CIFAR-10 datasets. 

\shortsection{Results on MINST dataset}
We can see from Figure \ref{fig4:subfig:1.a} and Figure \ref{fig4:subfig:1.b} that our proposed method can achieve $2.91\%$ test error when $\epsilon=7.0$, which is comparable to the $2.93\%$ test errors achieved by DP-SGD. Furthermore, the results show that our method is more efficient than DP-SGD in terms of iteration numbers and the training time. More specifically, our method is more than 2$\times$ faster than DP-SGD to achieve the desired test error.


\shortsection{Parameters for CNN5} We choose three different privacy budgets $\epsilon\in\{6.0,8.0,10.0\}$, and set $\delta=10^{-5}$. We set the clipping parameter $C_1=2$ for the term $\|\nabla f_i(\theta^t)\|_2$. For the term $\|\nabla f_i(\theta^t)-\nabla f_i(\theta^{t-1})\|_2$, we choose the clipping parameter $C_2$ by searching the grid $\{0.01,0.1,0.3,0.5,0.7,0.9,0.99\}$. 
 For DP-SGD, we tune the batch size by searching the grid $\{32, 64, 128\}$ and the step size by $\{0.01,0.02,0.05,0.1,0.2\}$. For DP-SRM, we tune the batch size $b$ by searching the grid $\{32, 64, 128\}$, step size by $\{0.01,0.02,0.05,0.1,0.2\}$, and $b_0$ by $\{b,2b,4b\}$. In addition, we set the momentum parameter $\gamma=C_2$.

\shortsection{Results for CNN5 on CIFAR-10 dataset} Figures \ref{fig4:subfig:1.e}-\ref{fig6:subfig:1.j} present the average test error of different methods versus the number of iterations as well as the training time under different privacy budgets for CNN5 on CIFAR-10 dataset. The CNN5 trained by the non-private SGD will have $39.5\%$ test error after 100  epochs. The results show that that our proposed method has $50.3\%$, $48.2\%$ and $47.1\%$ test errors when $\epsilon=6.0$, $\epsilon=8.0$ and  $\epsilon=10.0$. Nevertheless, DP-SGD has $51.0\%$, $50.2\%$ and $49.3\%$ test errors under the privacy budgets $\epsilon=6.0$, $\epsilon=8.0$ and  $\epsilon=10.0$, which are worse than our method. Furthermore, we can see from the plots that compared with DP-SGD, our method can reduce both the iteration numbers and the training time.

\shortsection{Results for CNN6 on CIFAR-10 dataset} Figure \ref{fig4:subfig:1.c} and Figure \ref{fig4:subfig:1.d} illustrate the average test error of different methods versus the number of iterations and the training time for CNN6 on CIFAR-10 dataset. We can see from the results that that our proposed method can achieve $29.3\%$ test errors given the privacy budget $\epsilon=8.0$, which are comparable to the results of DP-SGD with  $29.4\%$ under the same privacy budget. However, we can see from the plots that our method can significantly reduce the iteration numbers and the training time. When $\epsilon=8$, DP-SGD takes $5.8\times 10^{4}$ iterations and 5176 seconds to achiever $29.4\%$ test error. In sharp contrast, our method only takes $2.6\times 10^{4}$ iterations and 2589 seconds to achieve $29.3\%$ test error. 

\section{Proof of Main Results}\label{proof}
In this section, we present the proofs of our main results.
\subsection{Proof of Theorem \ref{thm:dp}}
We will provide the privacy guarantee of Algorithm \ref{alg:DPSRM} in this subsection. To this end, we need the following composition rule for RDP.
\begin{lemma}[\citet{mironov2017renyi}]\label{lemma:com_post}
	If $k$ randomized mechanisms $\cM_i:\cS^n\rightarrow\cR$ for $i\in[k]$, satisfy $(\alpha,\rho_i)$-RDP, then their composition $\big(\cM_1(S),\ldots,\cM_k(S)\big)$ satisfies $(\alpha,\sum_{i=1}^k\rho_i)$-RDP. Moreover, the input of the $i$-th mechanism can base on the outputs of previous $(i-1)$ mechanisms.
\end{lemma}

We will first show that our proposed algorithm satisfies RDP using Lemma \ref{lemma:GaussianM_RDP} and Lemma \ref{lemma:com_post}. Then we will transform it into $(\epsilon,\delta)$-DP based on Lemma \ref{lemma:RDP_to_DP}. For the given dataset $S$, we use $S^\prime$ to denote its neighboring dataset with one different example indexed by $i^\prime$ in the following discussion. According to Algorithm \ref{alg:DPSRM}, we use the following $\cM_t$ to denote the mechanism at $t$-th iteration
\begin{align}\label{eq:mechanismM}
    \cM_{t}=
    \left\{
	\begin{array} {ll}
		\nabla F_{\cB_t}(\btheta^t)+(1-\gamma)\big(\vb^{t-1}_p-\nabla F_{\cB_t}(\btheta^{t-1})\big)+\ub^t, & t>0,\\
		\vb^{0}+\ub^0, &  t=0.
	\end{array}
	\right.
\end{align}

 Therefore, our goal is to show the privacy guarantees of $\cM_t$ for $t=0,1,\ldots,T$.

\noindent\textbf{Case 1:} If $t=0$, we have $\vb^{0}=\nabla F_{\cB_0}(\btheta^0)$ and $\cM_0$ is equivalent to the following Gaussian mechanism 
\begin{align*}
    \cG_{0}=\nabla F_{\cB_0}(\btheta^0)+\ub^{0},
\end{align*}
where $\ub^{0}\sim N(0,\sigma^2_{0}\Ib_d)$.  Note that the mechanism $\cG_0$ is based on the subsampling, thus we will use the results of privacy-amplification by subsampling, i.e., Lemma \ref{lemma:GaussianM_RDP}, to show that $\cG_0$ satisfies RDP given appropriate $\ub^0$. To this end, we first consider the following Gaussian mechanism without subsampling
\begin{align*}
    \tilde \cG_{0}=\frac{1}{b_0}\sum_{i=1}^n\nabla f_i(\btheta^0)+\ub^{0}.
\end{align*}

\noindent\textbf{Sensitivity.} Consider the query on the dataset $S$ as follows $\tilde \qb_{0}(S)=\sum_{i=1}^n\nabla f_i(\btheta^0)/b_0$, where $\tilde \qb_{0}(S)$ denotes that the query is based on the dataset $S$. Thus, we have 
\begin{align*}
    \tilde \qb_{0}(S)-\tilde \qb_{0}(S^\prime)=\frac{1}{b_0}\big(\nabla f_i(\btheta^{0})-\nabla f_{i^\prime}(\btheta^{0})\big).
\end{align*}
Since each component function is $G$-Lipschitz, we can obtain the $\ell_2$-sensitivity of this query as follows
\begin{align}\label{eq:senstivity_gamma_t_1}
    \tilde \Delta_0=\frac{1}{b_0}\|\nabla f_i(\btheta^{0})-\nabla f_{i^\prime}(\btheta^{0})\|_2\leq \frac{2G}{b_0}.
\end{align}

\noindent\textbf{Privacy guarantee of $\cG_{0}$.}
By Lemma \ref{lemma:GaussianM_RDP}, if the Gaussian noise $\ub^0$ in $\tilde \cG_0$ has the following variance
\begin{align}\label{eq:sigma_1_RDP}
    \sigma^2_{0}=\frac{14T\alpha G^2}{\beta n^{2}\epsilon},
\end{align}
the mechanism $\tilde \cG_{0}$ satisfies $\big(\alpha,\beta\epsilon n^2/\big(7b_0^2T\big)\big)$-RDP. Therefore, according to the privacy-amplification by subsampling result in Lemma \ref{lemma:GaussianM_RDP}, we have that the mechanism $\cG_0$ satisfies $(\alpha,\rho_0)$-RDP, where $\rho_0=\beta \epsilon/T$. Furthermore, the variance $\sigma_0^2$ should satisfy the following condition
\begin{align*}
    \frac{\sigma_0^2}{\tilde \Delta_0^2}=\frac{\sigma_0^2b_0^2}{4G^2}=\frac{7b_0^2T\alpha}{\beta n^2\epsilon}\geq 0.7.
\end{align*}
And the parameter $\alpha$ should satisfy $\alpha\leq1+ 2(\sigma_0/\tilde \Delta_0)^2\log\big(1/\tau\alpha (1+(\sigma_0/\tilde \Delta_0)^2)\big)/3$.

\noindent\textbf{Case 2:} If $t>0$, according to the definition of $\cM_t$ in \eqref{eq:mechanismM}, we consider the following Gaussian mechanism
\begin{align*}
    \cG_t=\nabla F_{\cB_t}(\btheta^t)-(1-\gamma)\nabla F_{\cB_t}(\btheta^{t-1})+\ub^t.
\end{align*}
Now, we are going to show that $\cG_t$ satisfies RDP given appropriate $\ub^t$. Since the mechanism $\cG_t$ is based on the subsampling, we will use the similar proof procedure as in \textbf{Case 1} to show that $\cG_t$ satisfies RDP. Thus we consider the following Gaussian mechanism without subsampling
\begin{align*}
    \tilde \cG_t=\frac{1}{b}\sum_{i=1}^n\nabla f_i(\btheta^t)-(1-\gamma)\frac{1}{b}\sum_{i=1}^n\nabla f_i(\btheta^{t-1})+\ub^t.
\end{align*}

\noindent\textbf{Sensitivity.} We consider the following query without subsampling
\begin{align*}
    \tilde \qb_t(S)=\frac{1}{b}\sum_{i=1}^n\nabla f_i(\btheta^t)-(1-\gamma)\frac{1}{b}\sum_{i=1}^n\nabla f_i(\btheta^{t-1}).
\end{align*}
Thus we have
 \begin{align*}
     \tilde \qb_t(S)-\tilde \qb_t(S^\prime)&=\frac{1}{b}\big(\nabla f_i (\btheta^t)-(1-\gamma)\nabla f_i (\btheta^{t-1})-\nabla f_{i^\prime} (\btheta^t)+(1-\gamma)\nabla f_{i^\prime}(\btheta^{t-1})\big).
 \end{align*}
 As a result, we can obtain the $\ell_2$-sensitivity of the query $\tilde \qb_t$ as follows
 \begin{align*}
     \tilde \Delta_t&=\frac{1}{b}\big\|(1-\gamma)\big(\nabla f_i (\btheta^t)-\nabla f_i (\btheta^{t-1})-\nabla f_{i^\prime} (\btheta^t)+\nabla f_{i^\prime}(\btheta^{t-1})\big)\\
     &~~\qquad+\gamma\big(\nabla f_i (\btheta^{t})-\nabla f_{i^\prime} (\btheta^{t})\big) \big\|_2\\
     &\leq\frac{2L(1-\gamma)}{b}\|\btheta^t-\btheta^{t-1}\|_2+\frac{2\gamma G}{b},
 \end{align*}
where the inequality is due to $L$-Lipschitz continuous gradient and $G$-Lipschitz of each component function. Furthermore, according to the update rule of Algorithm \ref{alg:DPSRM} and the definition of $\eta_{t-1}$, we have
\begin{align*}
    \|\btheta^t-\btheta^{t-1}\|_2\leq \eta_{t-1}\|\vb^{t-1}_p\|_2\leq \min\bigg\{\frac{\zeta}{n_0L\|\vb^{t-1}_p\|_2},\frac{1}{2n_0L}\bigg\}\cdot\|\vb^{t-1}_p\|_2\leq \frac{\zeta}{n_0L},
\end{align*}
which implies that
\begin{align}\label{eq:senstivity_gamma_t}
   \tilde \Delta_t\leq\frac{2L(1-\gamma)}{b}\|\btheta^t-\btheta^{t-1}\|_2+\frac{2\gamma G}{b}\leq \frac{2\big((1-\gamma)\zeta/n_0+\gamma G\big)}{b}.
\end{align}

\noindent\textbf{Privacy guarantee of $\cG_t$.}
By Lemma \ref{lemma:GaussianM_RDP}, if we the Gaussian noise $\ub^t$ in $\tilde \cG_t$ has the variance as follows
\begin{align}\label{eq:sigma_2_RDP}
    \sigma^2_t&=\frac{14T\alpha\big((1-\gamma)\zeta/n_0+\gamma G\big)^2}{\beta n^2\epsilon},
\end{align}
the mechanism $\tilde \cG_t$ satisfies $\big(\alpha,\beta\epsilon n^2/\big(7b^2T\big)\big)$-RDP. Thus based on the privacy-amplification by subsampling result (Lemma \ref{lemma:GaussianM_RDP}), we can get that the mechanism $\cG_t$ satisfies $(\alpha,\rho)$-RDP, where $\rho=\beta \epsilon/T$. In addition, the variance $\sigma_t^2$ should satisfy the following condition 
\begin{align*}
    \frac{\sigma_t^2}{\tilde \Delta_t^2}=\frac{\sigma_t^2b^2}{4\big((1-\gamma)\zeta/n_0+\gamma G\big)^2}=\frac{7b^2T\alpha}{\beta n^2\epsilon}\geq 0.7.
\end{align*}
And the parameter $\alpha$ should satisfy $\alpha\leq1+ 2(\sigma_t/\tilde \Delta_t)^2\log\big(1/\tau\alpha (1+(\sigma_t/\tilde \Delta_t)^2)\big)/3$. As a result, we show that $\cG_t$ satisfies $(\alpha,\rho)$-RDP.

\noindent\textbf{Privacy guarantee of $\cM_{t}$.}
By the definition of the mechanism $\cM_{t}$ in \eqref{eq:mechanismM}, $\cM_t$ is a composition of $\cG_{0},\ldots,\cG_{t}$, i.e., $\cM_{t}=(\cG_{0},\ldots,\cG_{t})$. According to the composition property of RDP, i.e., Lemma \ref{lemma:com_post}, we have 
$\cM_{t}$ satisfies $(\alpha,\rho_0+(t-1)\rho)$-RDP. Since $\rho_0=\rho=\beta\epsilon/T$, we have that after $T^\prime$ iterations of Algorithm \ref{alg:DPSRM}, it satisfies $(\alpha,\beta T^\prime\epsilon/T)$-RDP.  According to Lemma \ref{lemma:RDP_to_DP} and $\alpha=\log(1/\delta)/\big((1-\beta)\epsilon\big)+1$, we have that after $T^\prime$ iterations, Algorithm \ref{alg:DPSRM} satisfies $(T^\prime\epsilon/T,\delta)$-DP. As a result, we have that for each $\btheta^t$, where $t=1,\ldots,T$, it satisfies $(\epsilon,\delta)$-DP. Finally, by the definition of $\tilde \btheta$, we have $\tilde \btheta$ satisfies $(\epsilon,\delta)$-DP.

\subsection{Proof of Corollary \ref{coro:dp}}
In this subsection, we show that by choosing a larger mini-batch size, we can get rid of the constraints in Theorem \ref{thm:dp}. More specifically, let $b_0^2=b^2= n^2\epsilon/T$ and $\beta=1/2$, we have $\sigma^{\prime 2}=7T\alpha b^2/(\beta n^2\epsilon)=14\alpha$. Furthermore, we have
\begin{align*}
   \tau\alpha\big(1+\sigma^{\prime2}) \stackrel{\mathrm{(i)}}\leq 15\tau\alpha^2 \stackrel{\mathrm{(ii)}}=15\big(2\log(1/\delta)/\epsilon+1\big)^2\sqrt{\epsilon/T},
\end{align*}
where $\mathrm{(i)}$ uses $\sigma^{\prime 2}=14\alpha$, $\mathrm{(ii)}$ uses $\tau=b/n=\sqrt{\epsilon/T}$ and $\epsilon=2\log(1/\delta)/(\alpha-1)$.
If $\epsilon\leq 2\log(1/\delta)$, we can obtain $\tau\alpha\big(1+\sigma^{\prime2})\leq 1/3$ if $T$ is larger than $O\big(\log^4(1/\delta)/\epsilon^3\big)$. If $\epsilon> 2\log(1/\delta)$, we can obtain $\tau\alpha\big(1+\sigma^{\prime2})\leq 1/3$ if $T$ is larger than $O(\epsilon)$. Therefore, we can get $\log(1/\tau\alpha\big(1+\sigma^{\prime2})\big)\geq 1$. As a result, we have $2\big(\sigma^{\prime2}\log(1/\tau\alpha\big(1+\sigma^{\prime2})\big)\big)/3\geq 28\alpha/3 >\alpha-1$.

\subsection{Proof of Theorem \ref{thm:utility}}
In this subsection. we provide the utility guarantee of our method. According to the assumption that each component function has $L$-Lipschitz continuous gradient, we can obtain that
\begin{align*}
    \|\nabla F(\xb)-\nabla F(\yb)\|_2=\frac{1}{n}\sum_{i=1}^n\|\nabla f_i(\xb)-\nabla f_i(\yb)\|_2\leq L\|\xb-\yb\|_2,
\end{align*}
which implies that $F(\xb)$ has $L$-Lipschitz continuous gradient. Thus we have
\begin{align*}
    F(\btheta^{t+1})&\leq F(\btheta^t)+\la\nabla F(\btheta^t),\btheta^{t+1}-\btheta^{t}\ra+\frac{L}{2}\|\btheta^{t+1}-\btheta^{t}\|_2^2\\
    &=F(\btheta^t)-\eta_t\la\nabla F(\btheta^t),\vb^t_p\ra+\frac{\eta_t^2L}{2}\big\|\vb^t_p\big\|_2^2\\
    &=F(\btheta^t)+\frac{\eta_t}{2}\big\|\nabla F(\btheta^t)-\vb^t_p\big\|_2^2-\frac{\eta_t}{2}\big\|\nabla F(\btheta^t)\big\|_2^2-\eta_t\bigg(\frac{1}{2}-\frac{\eta_tL}{2}\bigg)\big\|\vb^t_p\big\|_2^2,
\end{align*}
where the last equality is due to the fact that $2\la\nabla F(\btheta^t),\vb^t_p\ra=\big\|\nabla F(\btheta^t)\big\|_2^2+\big\|\vb^t_p\big\|_2^2-\big\|\nabla F(\btheta^t)-\vb^t_p\big\|_2^2$. Since $\eta_t\leq 1/(2n_0L)$, we can obtain that 
\begin{align*}
    F(\btheta^{t+1})&\leq F(\btheta^t)+\frac{1}{4n_0L}\big\|\nabla F(\btheta^t)-\vb^t_p\big\|_2^2-\frac{\eta_t}{4}\big\|\vb^t_p\big\|_2^2.
\end{align*}
 In addition, we have
\begin{align*}
    \frac{\eta_t}{4}\big\|\vb^t_p\big\|_2^2=\frac{\zeta^2}{8n_0L}\min\big\{2\big\|\vb^t_p/\zeta\big\|_2,\big\|\vb^t_p/\zeta\big\|_2^2\big\}\geq \frac{\zeta\big\|\vb^t_p\big\|_2-2\zeta^2}{4n_0L}.
\end{align*}
Thus we have
\begin{align}\label{eq:eq0}
    F(\btheta^{t+1})&\leq F(\btheta^t)+\frac{1}{4n_0L}\big\|\nabla F(\btheta^t)-\vb^t_p\big\|_2^2-\frac{\zeta\big\|\vb^t_p\big\|_2}{4n_0L}+\frac{\zeta^2}{2n_0L}.
\end{align}
Summing over $t=0,\ldots,T-1$ and taking expectation in \eqref{eq:eq0}, we can get
\begin{align}\label{eq:contraction}
    \frac{\zeta}{4n_0L}\sum_{t=0}^{T-1}\EE\big\|\vb^t_p\big\|_2&\leq F(\btheta^0)-\EE F(\btheta^T)+\frac{1}{4n_0L}\sum_{t=0}^{T-1}\EE\big\|\nabla F(\btheta^t)-\vb^t_p\big\|_2^2+\frac{T\zeta^2}{2n_0L}\nonumber\\
    &\leq F(\btheta^0)- F(\btheta^*)+\frac{1}{4n_0L}\sum_{t=0}^{T-1}\EE\big\|\nabla F(\btheta^t)-\vb^t_p\big\|_2^2+\frac{T\zeta^2}{2n_0L}.
\end{align}
For the term $\EE\big\|\nabla F(\btheta^t)-\vb^t_p\big\|_2^2$, we can bound it as follows: we first consider the conditional expectation
\begin{align}\label{eq:eq1}
    \EE_{t}\big\|\vb^t_p-\nabla F(\btheta^t)\big\|_2^2&=\EE_{t}\big\|(1-\gamma)\big(\vb^{t-1}_p-\nabla F_{\cB_t}(\btheta^{t-1})\big)+\nabla F_{\cB_t}(\btheta^t)-\nabla F(\btheta^t)+\ub^t\big\|_2^2\nonumber\\
    &=\EE_{t}\big\|(1-\gamma)\big(\vb^{t-1}_p-\nabla F(\btheta^{t-1})\big)+(1-\gamma)\nabla F(\btheta^{t-1})\nonumber\\
    &\quad\qquad-(1-\gamma)\nabla F_{\cB_t}(\btheta^{t-1})+\nabla F_{\cB_t}(\btheta^t)-\nabla F(\btheta^t)\big\|_2^2+\EE_{t}\|\ub^t\|_2^2\nonumber\\
    &=\EE_{t}\big\|(1-\gamma)\big(\vb^{t-1}_p-\nabla F(\btheta^{t-1})\big)+(1-\gamma)\big(\nabla F_{\cB_t}(\btheta^t)-\nabla F_{\cB_t}(\btheta^{t-1})\nonumber\\
    &~~\qquad+\nabla F(\btheta^{t-1})-\nabla F(\btheta^t)\big)
    +\gamma\big(\nabla F_{\cB_t}(\btheta^t)-\nabla F(\btheta^t)\big)\big\|_2^2+\EE_{t}\|\ub^t\|_2^2,
\end{align}
where $\EE_t$ is taken over the randomness at the $t$-th iteration given the observations after $(t-1)$-th iteration, the first equation comes from the definition of $\vb^t_p$, the second one is due to the independence of the random variables. Therefore, we can obtain that 
\begin{align}\label{eq:eq2}
    \EE_{t}\big\|\vb^t_p-\nabla F(\btheta^t)\big\|_2^2&=(1-\gamma)^2\EE_{t}\big\|\vb^{t-1}_p-\nabla F(\btheta^{t-1})\big\|_2^2\nonumber\\
    &\qquad+2\gamma^2\EE_{t}\big\|\nabla F_{\cB_t}(\btheta^t)-\nabla F(\btheta^t)\big\|_2^2+\EE_{t}\|\ub^t\|_2^2\nonumber\\
    &\qquad+2(1-\gamma)^2\EE_{t}\big\|\nabla F_{\cB_t}(\btheta^t)-\nabla F_{\cB_t}(\btheta^{t-1})+\nabla F(\btheta^{t-1})-\nabla F(\btheta^t)\big\|_2^2,
\end{align}
where the equality is due to the expansion of \eqref{eq:eq1} and Cauchy-Schwartz inequality.
In addition, we have
\begin{align*}
    &\EE_{t}\big\|\nabla F(\btheta^t)-\nabla F(\btheta^{t-1})-\nabla F_{\cB_t}(\btheta^t)+\nabla F_{\cB_t}(\btheta^{t-1})\big\|_2^2\nonumber\\
    &\leq \frac{1}{b}\cdot\frac{1}{n}\sum_{i=1}^n\big\|\nabla F(\btheta^t)-\nabla F(\btheta^{t-1})-\nabla f_i(\btheta^t)+\nabla f_i(\btheta^{t-1})\big\|_2^2\nonumber\\
    &\leq\frac{1}{b}\cdot\frac{1}{n}\sum_{i=1}^n\big\|\nabla f_i(\btheta^t)-\nabla f_i(\btheta^{t-1})\big\|_2^2\nonumber\\
    &\leq \frac{L^2}{b}\|\btheta^t-\btheta^{t-1}\|_2^2,
\end{align*}
where the first inequality is due to Lemma \ref{lemma:scsg}, the second one comes from the fact that $\EE\|\bX-\EE\bX\|_2^2\leq \EE\|\bX\|_2^2$ for any random variable $\bX$, and the last one is due to the gradient Lipschitz property of each component function. According to the update rule, we have
\begin{align*}
    \|\btheta^t-\btheta^{t-1}\|_2\leq \eta_{t-1}\big\|\vb^{t-1}_p\big\|_2\leq \min\bigg\{\frac{\zeta}{n_0L\big\|\vb^{t-1}_p\big\|_2},\frac{1}{2n_0L}\bigg\}\cdot\big\|\vb^{t-1}_p\big\|_2\leq \frac{\zeta}{n_0L},
\end{align*}
which implies 
\begin{align}\label{eq:eq3}
    \EE_{t}\big\|\nabla F(\btheta^t)-\nabla F(\btheta^{t-1})-\nabla F_{\cB_t}(\btheta^t)+\nabla F_{\cB_t}(\btheta^{t-1})\big\|_2^2\leq \frac{\zeta^2}{n_0^2b}.
\end{align}

Thus plugging \eqref{eq:eq3} into \eqref{eq:eq2}, we can obtain that
\begin{align}\label{eq:eq4}
    \EE_{t}\big\|\vb^t_p-\nabla F(\btheta^t)\big\|_2^2&\leq (1-\gamma)^2\big\|\vb^{t-1}_p-\nabla F(\btheta^{t-1})\big\|_2^2+\frac{2(1-\gamma)^2L^2}{b}\|\btheta^t-\btheta^{t-1}\|_2^2\nonumber\\
    &\qquad+2\gamma^2\EE_{t}\big\|\nabla F_{\cB_t}(\btheta^t)-\nabla F(\btheta^t)\big\|_2^2+\EE_{t}\|\ub^t\|_2^2\nonumber\\
    &\leq (1-\gamma)^2\big\|\vb^{t-1}_p-\nabla F(\btheta^{t-1})\big\|_2^2+\frac{2(1-\gamma)^2\zeta^2}{n_0^2b}+\frac{2\gamma^2G^2}{b}+\EE_{t}\|\ub^t\|_2^2,
\end{align}
where the second inequality follows the following inequality (using Lemma \ref{lemma:scsg}, $\EE\|\bX-\EE\bX\|_2^2\leq \EE\|\bX\|_2^2$, and the $G$-Lipschitz of each component function)
\begin{align}\label{eq:varaince1}
    \EE_{t}\big\|\nabla F_{\cB_t}(\btheta^t)-\nabla F(\btheta^t)\big\|_2^2\leq \frac{1}{b}\cdot\frac{1}{n}\sum_{i=1}^n\big\|\nabla f_i(\btheta^t)\big\|_2^2\leq \frac{G^2}{b}.
\end{align}
Therefore, taking expectations over all iterations in \eqref{eq:eq4}, we can get
\begin{align}\label{eq:eq5}
     \EE\big\|\vb^t_p-\nabla F(\btheta^t)\big\|_2^2&\leq (1-\gamma)^2\EE\big\|\vb^{t-1}_p-\nabla F(\btheta^{t-1})\big\|_2^2+\frac{2(1-\gamma)^2\zeta^2}{n_0^2b}+\frac{2\gamma^2G^2}{b}+d\sigma^2.
\end{align}
Following the proof of Lemma 9 in \cite{yuan2020stochastic}, we have
\begin{align*}
    \gamma\sum_{t=0}^{T-1}\EE\big\|\vb^t_p-\nabla F(\btheta^t)\big\|_2^2&\leq \frac{2T(1-\gamma)^2\zeta^2}{n_0^2b}+\frac{2T\gamma^2G^2}{b}+Td\sigma^2+\EE\big\|\vb^0_p-\nabla F(\btheta^0)\big\|_2^2\\
    &\leq \frac{2T(1-\gamma)^2\zeta^2}{n_0^2b}+\frac{2T\gamma^2G^2}{b}+Td\sigma^2+\frac{G^2}{b_0}+d\sigma_0^2,
\end{align*}
where the last line comes from the definition of $\vb_p^0=\nabla F_{\cB_0}(\btheta^0)+\ub^0$ and the inequality $ \EE\big\|\nabla F_{\cB_0}(\btheta^0)-\nabla F(\btheta^0)\big\|_2^2\leq G^2/b_0$ (see equation \eqref{eq:varaince1}).
Therefore, we can obtain that 
\begin{align}\label{eq:eq6}
    \sum_{t=0}^{T-1}\EE\big\|\vb^t_p-\nabla F(\btheta^t)\big\|_2^2\leq \frac{2T(1-\gamma)^2\zeta^2}{n_0^2\gamma b}+\frac{2T\gamma G^2}{b}+\frac{Td\sigma^2+d\sigma_0^2}{\gamma}+\frac{G^2}{\gamma b_0}.
\end{align}

Combining \eqref{eq:contraction} and \eqref{eq:eq6}, we can get
\begin{align*}
    \frac{\zeta}{4n_0L}\sum_{t=0}^{T-1}\EE\big\|\vb^t_p\big\|_2
    &\leq F(\btheta^0)- F(\btheta^*)+\frac{1}{4n_0L}\sum_{t=0}^{T-1}\EE\big\|\nabla F(\btheta^t)-\vb^t_p\big\|_2^2+\frac{T\zeta^2}{2n_0L}\\
    &\leq  F(\btheta^0)- F(\btheta^*)+ \frac{T(1-\gamma)^2\zeta^2}{2n_0^3L\gamma b}+\frac{T\gamma G^2}{ 4Ln_0b}\nonumber\\
    &\qquad+\frac{Td\sigma^2+d\sigma_0^2}{4n_0L\gamma}+\frac{G^2}{4L\gamma n_0b_0}+\frac{T\zeta^2}{2n_0L}.
\end{align*}
Hence we have 
\begin{align}\label{eq:eq7}
   \frac{1}{T}\sum_{t=0}^{T-1}\EE\big\|\vb^t_p\big\|_2
    &\leq \frac{4n_0L}{T\zeta}\big(F(\btheta^0)-F(\btheta^*)\big)+ \frac{2\zeta}{n_0^2\gamma b}+\frac{\gamma G^2}{ \zeta b}+\frac{d\sigma^2+d\sigma_0^2/T}{\zeta\gamma}+\frac{G^2}{T\zeta\gamma b_0}+2\zeta\nonumber\\
    &\leq 6\zeta+\frac{2\zeta}{n_0^2\gamma b}+\frac{\gamma G^2}{ \zeta b}+\frac{d\sigma^2+d\sigma_0^2/T}{\zeta\gamma}+\frac{G^2}{T\zeta\gamma b_0},
\end{align}
where the first inequality is due to $T= \lfloor 4n_0L\big(F(\btheta^0)-F(\btheta^*)\big)/\zeta^2\rfloor +1$. In addition, according to \eqref{eq:eq6} and Jensen's inequality, we have
\begin{align}\label{eq:eq8}
    \frac{1}{T}\sum_{t=0}^{T-1}\EE\big\|\nabla F(\btheta^t)-\vb^t_p\big\|_2\leq \frac{\sqrt{2}\zeta}{n_0\sqrt{\gamma b}}+\frac{\sqrt{2\gamma} G}{\sqrt{b}}+\frac{\sqrt{d}\sigma+\sqrt{d}\sigma_0/\sqrt{T}}{\sqrt{\gamma}}+\frac{G}{\sqrt{T\gamma b_0}}.
\end{align}
Thus by the definition of $\tilde \btheta$, we have 
\begin{align}\label{eq:eq9}
    \EE\|\nabla F(\tilde \btheta)\|_2&=\frac{1}{T}\sum_{t=0}^{T-1}\EE\|\nabla F(\btheta^t)\|_2\nonumber\\
    &\leq \frac{1}{T}\sum_{t=0}^{T-1}\EE\big\|\vb^t_p\big\|_2+\frac{1}{T}\sum_{t=0}^{T-1}\EE\big\|\nabla F(\btheta^t)-\vb^t_p\big\|_2\nonumber\\
    &\leq 6\zeta+\frac{2\zeta}{n_0^2\gamma b}+\frac{\gamma G^2}{ \zeta b}+\frac{d\sigma^2}{\zeta\gamma}+\frac{d\sigma_0^2}{T\zeta\gamma}+\frac{G^2}{T\zeta\gamma b_0}+\frac{\sqrt{2}\zeta}{n_0\sqrt{\gamma b}}+\frac{\sqrt{2\gamma} G}{\sqrt{b}}\nonumber\\
    &\qquad+\frac{\sqrt{d}\sigma}{\sqrt{\gamma}}+\frac{\sqrt{d}\sigma_0}{\sqrt{T\gamma}}+\frac{G}{\sqrt{T\gamma b_0}},
\end{align}
where the second inequality comes from \eqref{eq:eq7} and \eqref{eq:eq8}. Let $\gamma^2=2\zeta^2/(n_0^2G^2)$, $b=G/(n_0\zeta)$, $b_0=G^3/(\zeta LD_F)$, where $D_F=F(\btheta^0)-F(\btheta^*)$ and $F(\btheta^*$) is a global minimum of $F$, by the definition of $T$, we can get
\begin{align}\label{eq:utility_single}
    \EE\|\nabla F(\tilde \btheta)\|_2\leq 15\zeta+\frac{d\sigma^2}{\zeta\gamma}+\frac{\sqrt{d}\sigma}{\sqrt{\gamma}}+\frac{d\sigma_0^2}{T\zeta\gamma}+\frac{\sqrt{d}\sigma_0}{\sqrt{T\gamma}}.
\end{align}
 Furthermore, we have 
\begin{align}\label{eq:eq9}
    \sigma^2=\frac{14T\big((1-\gamma)\zeta/n_0+\gamma G\big)^2\log(1/\delta)}{n^{2}\epsilon^2},\quad \sigma_0^2=\frac{14TG^2\log(1/\delta)}{n^2\epsilon^2}.
\end{align}
Plugging \eqref{eq:eq9} into \eqref{eq:utility_single}, we can obtain 
\begin{align}\label{eq:final_bound}
    \EE\|\nabla F(\tilde \btheta)\|_2&\leq 15\zeta+\frac{C_1Td G\log(1/\delta)}{n_0n^2\epsilon^2}+\frac{\sqrt{C_1T\zeta dG\log(1/\delta)}}{n\epsilon\sqrt{n_0}}+\frac{C_2dn_0G^3\log(1/\delta)}{n^2\epsilon^2\zeta^2}\nonumber\\   &\qquad+\frac{\sqrt{C_2n_0dG^3\log(1/\delta)}}{n\epsilon\sqrt{\zeta}}\nonumber\\
    &\leq 15\zeta+\frac{C_3LD_FGd\log(1/\delta)}{n^2\epsilon^2\zeta^2}+\frac{\sqrt{C_4GLD_Fd\log(1/\delta)}}{n\epsilon\sqrt{\zeta}}\nonumber\\
    &\qquad+\frac{C_5n_0dG^3\log(1/\delta)}{n^2\epsilon^2\zeta^2}+\frac{\sqrt{C_6n_0dG^3\log(1/\delta)}}{n\epsilon\sqrt{\zeta}},
\end{align}
where the second inequality is due to the fact that $T= \lfloor 4n_0LD_F/\zeta^2\rfloor +1$. Without loss of generality, we can assume $G\geq 1$ and $\zeta\leq 1$.
Therefore, let $n_0=LD_F/G^2\cdot(G/\zeta)^\kappa$ with $\kappa\in[0,1]$, and plugging $n_0$ into \eqref{eq:final_bound}, we can obtain
\begin{align}\label{eq:final_bound_f1}
  \EE\|\nabla F(\tilde \btheta)\|_2&\leq 15\zeta+ \frac{C_7LD_FGd\log(1/\delta)G^\kappa}{n^2\epsilon^2\zeta^{2+\kappa}}+\frac{C_8\sqrt{GLD_Fd\log(1/\delta)}G^{\frac{\kappa}{2}}}{n\epsilon\zeta^{\frac{1+\kappa}{2}}}.
\end{align}
Thus, choosing 
\begin{align}\label{eq:kappa}
\zeta=C_9\bigg(\frac{G^{\frac{\kappa}{2}}\sqrt{GLD_Fd\log(1/\delta)}}{n\epsilon}\bigg)^{\frac{2}{3+\kappa}},
\end{align}
we can get
\begin{align}\label{eq:final_bound_f2}
  \EE\|\nabla F(\tilde \btheta)\|_2&\leq C_{10}\bigg(\frac{G^{\frac{\kappa}{2}}\sqrt{GLD_Fd\log(1/\delta)}}{n\epsilon}\bigg)^{\frac{2}{3+\kappa}}.
\end{align}
Note that we require $\gamma\leq 1$, which gives us $n\epsilon\geq O\big(G^2(d\log(1/\delta))^{1/2}/(LD_F)\big)$.

Furthermore, according to Theorem \ref{thm:dp}, to achieve the desired privacy guarantee, we require $\sigma^{\prime2}=\min\{b^2\sigma^2/\big(4((1-\gamma)\zeta/n_0+\gamma G)^2\big), b_0^2\sigma_0^2/(4G^2)\}\geq 0.7$. Note that $b=G/(n_0\zeta)$, $b_0=G^3/(\zeta LD_F)$, $n_0=LD_F/G^2\cdot(G/\zeta)^\kappa$, we have $b=b_0\cdot(\zeta/G)^\kappa$. Thus, the aforementioned requirement reduces to 
\begin{align*}
    \frac{14b_0^2T\log(1/\delta)}{4n^2\epsilon^2}\cdot\frac{\zeta^{2\kappa}}{G^{2\kappa}}&=\frac{14b_0^2n_0LD_{F}\log(1/\delta)}{\zeta^4n^2\epsilon^2}\cdot\frac{\zeta^{2\kappa}}{G^{2\kappa}}\\
    &\geq \frac{14b_0n_0LD_{F}\log(1/\delta)}{\zeta^4n^2\epsilon^2}\cdot\frac{\zeta^{2\kappa}}{G^{2\kappa}}\\
    &=\frac{14GLD_{F}\log(1/\delta)}{\zeta^3n^2\epsilon^2}\cdot\frac{\zeta^{\kappa}}{G^{\kappa}}\\
    &\geq 0.7,
\end{align*}
where the first equality comes from the definition of $T$ and the first inequality is due to $b_0\geq 1$.  Therefore, we need
\begin{align}\label{eq:kappa_r}
    \zeta\leq \bigg(4\frac{G^{-\frac{\kappa}{2}}\sqrt{GLD_Fd\log(1/\delta)}}{n\epsilon}\bigg)^{\frac{2}{3-\kappa}}.
\end{align}
Combining \eqref{eq:kappa} and \eqref{eq:kappa_r}, we need to choose $\kappa=0$ in $n_0$, which gives us 
\begin{align}\label{eq:final_bound_f3}
  \EE\|\nabla F(\tilde \btheta)\|_2&\leq C_{10}\bigg(\frac{\sqrt{GLD_Fd\log(1/\delta)}}{n\epsilon}\bigg)^{\frac{2}{3}},
\end{align}
where $\{C_i\}_{i=1}^{10}$ are absolute constants. Furthermore, the requirement 
$\alpha-1=\log(1/\delta)/\big((1-\beta)\epsilon\big)\leq 2\sigma^{\prime 2}\log\big(1/\big(\tau\alpha (1+\sigma^{\prime 2})\big)\big)/3$ in Theorem \ref{thm:dp} can be satisfied under our choice of parameters given large enough $n$. Since we have $\sigma^{\prime 2}\geq 0.7$, we have $2\sigma^{\prime 2}\log\big(1/\big(\tau\alpha (1+\sigma^{\prime 2})\big)\big)/3\geq 0.4\log\big(1/\big(\tau\alpha (1+\sigma^{\prime 2})\big)\big)\geq0.4\log\big(1/\big(3\tau\alpha \sigma^{\prime 2}\big)\big)$. Furthermore, we have
\begin{align*}
   \tau\alpha \sigma^{\prime 2}&=\frac{G^3}{n\zeta LD_F}\cdot\frac{\log(1/\delta)+(1-\beta)\epsilon}{(1-\beta)\epsilon}\cdot\frac{14GLD_{F}\log(1/\delta)}{\zeta^3n^2\epsilon^2}\\
   &\leq \frac{28G^4\log^2(1/\delta)}{(1-\beta)n^3\epsilon^3\zeta^4}\\
   &\leq C_{11} \frac{G^4\log^2(1/\delta)}{(n\epsilon)^3}\cdot\frac{(n\epsilon)^{8/3}}{(GLD_Fd\log(1/\delta))^{4/3}}\\
   &=C_{11}\frac{G^{8/3}\log^{2/3}(1/\delta)}{(n\epsilon)^{1/3}(LD_Fd)^{4/3}},
\end{align*}
where the first inequality comes from assumining $\epsilon\leq \log(1/\delta)$ without loss of generality, and the second inequality is due to the definition of $\zeta$. Thus we have
\begin{align*}
    \log\big(1/\big(3\tau\alpha \sigma^{\prime 2}\big)\big)\geq\log\bigg(3C_{11}\frac{(n\epsilon)^{1/3}(LD_Fd)^{4/3}}{G^{8/3}\log^{2/3}(1/\delta)}\bigg).
\end{align*}
As a result, the requirement reduces to 
\begin{align*}
    0.4\log\bigg(3C_{11}\frac{(n\epsilon)^{1/3}(LD_Fd)^{4/3}}{G^{8/3}\log^{2/3}(1/\delta)}\bigg)\geq \frac{\log(1/\delta)}{(1-\beta)\epsilon},
\end{align*}
which can be satisfied if we have 
\begin{align*}
    n\geq C_{12}\frac{G^{8}\log^{2}(1/\delta)}{(LD_Fd)^{4}\epsilon},
\end{align*}
where $C_{11},C_{12}$ are some large constants.

\textbf{Gradient Complexity.}
Since we have  $b=b_0=G^3/(\zeta LD_F)$, the total gradient complexity is 
\begin{align*}
    2(T-1)b+b_0\leq \frac{8LD_Fn_0}{\zeta^2}\cdot \frac{G^3}{LD_F\zeta}+\frac{G^3}{LD_F\zeta}.
\end{align*}
According to the definition of $\zeta$ and $n_0$, we have the total gradient complexity is $O\big(n^2\epsilon^2/(d\log(1/\delta))\big)$.


\section{Proof of Lemma \ref{lemma:GaussianM_RDP}}

Without loss of generality, we assume $\Delta(q)=1$. According to Theorem 9 in \citet{wang2018subsampled}, we have 
\begin{align}\label{eq:rho1}
    \rho^\prime(\alpha)\leq \frac{1}{\alpha-1}\log\bigg(1+\tau^2{\alpha \choose 2}\min\Big\{4(e^{\rho(2)}-1),2e^{\rho(2)}\Big\}+\sum_{j=3}^\alpha \tau^j{\alpha \choose j}2e^{(j-1)\rho(j)}\bigg),
\end{align}
where $\tau$ is the subsample rate, $\rho(j)=j/(2\sigma^2)$. Next, we will show that the summation term in the right hand side of the above inequality is dominated by the second term under certain conditions. First of all, when $\sigma^2$ is large, i.e., $\sigma^2\geq 0.7$, we have 
\begin{align*}
    \min\Big\{4(e^{\rho(2)}-1),2e^{\rho(2)}\Big\}\leq 6/\sigma^2,
\end{align*}
which implies that
\begin{align*}
    \tau^2{\alpha \choose 2}\min\Big\{4(e^{\rho(2)}-1),2e^{\rho(2)}\Big\}\leq \tau^2{\alpha \choose 2} 6/\sigma^2.
\end{align*}
Next, we consider the summation term in \eqref{eq:rho1}, and we have 
\begin{align*}
    \sum_{j=3}^\alpha \tau^j{\alpha \choose j}2e^{(j-1)\rho(j)}&\leq\tau^2{\alpha \choose 2}\bigg(\sum_{j=3}^\alpha\tau^{j-2}\alpha^{j-2}e^{\frac{(\alpha-1) j}{2\sigma^2}}\bigg)\\
    &\leq\tau^2{\alpha \choose 2}\frac{\tau\alpha e^{\frac{3(\alpha-1)}{2\sigma^2}}}{1-\tau\alpha e^{\frac{\alpha-1}{2\sigma^2}}},
\end{align*}
where the first inequality is due to the fact that
\begin{align*}
    e^{(j-1)\rho(j)}=e^{\frac{(j-1)j}{2\sigma^2}}\leq e^{\frac{(\alpha-1)j}{2\sigma^2}}\quad\text{and}\quad
     {\alpha \choose j }=\frac{\alpha!}{j!(\alpha-j)!}\leq \frac{\alpha^2\alpha^{j-2}}{3!}.
 \end{align*}
 In addition, the last inequality comes from the condition that $\tau\alpha\exp\big((\alpha-1)/(2\sigma^2)\big)<1$ and the sum of the geometric sequence.
Therefore, as long as 
\begin{align}\label{eq:alpha_ineq}
    \alpha-1\leq\frac{2}{3}\sigma^2\log\frac{1}{\tau\alpha(1+\sigma^2)},
\end{align}
we have 
\begin{align*}
    \sum_{j=3}^\alpha \tau^j{\alpha \choose j}2e^{(j-1)\rho(j)}\leq \tau^2{\alpha \choose 2}\frac{1}{\sigma^2}.
\end{align*}
In addition, we require that $\tau\alpha\exp\big((\alpha-1)/(2\sigma^2)\big)<1$. By plugging the condition of $\alpha$ into the above requirement, we can obtain that this condition can hold if $\tau<1$.

As a result, under the conditions that $\sigma^2\geq 0.7$, $\alpha\leq \log(1/\tau\big(1+\sigma^2)\big)$, we can obtain that
\begin{align*}
    \rho^\prime(\alpha)\leq \frac{1}{\alpha-1}\log\bigg(1+\tau^2{\alpha \choose 2}\frac{10}{\sigma^2}\bigg)\leq \frac{1}{\alpha-1}\tau^2{\alpha \choose 2}\frac{7}{\sigma^2}\leq 3.5\alpha\tau^2/\sigma^2.
\end{align*}

\section{Auxiliary Lemmas}
\begin{lemma}\citep{lei2017non}\label{lemma:scsg}
Consider vectors $\ab_i$ satisfying $\sum_{i=1}^n\ab_i=0$. Let $\cB$ be a uniform random subset of $\{1,2,\ldots,n\}$ with size $m$, we have
\begin{align*}
    \EE\bigg\|\frac{1}{m}\sum_{i\in\cB}\ab_i\bigg\|_2^2\leq\frac{\ind\{|\cB|< n \}}{mn}\sum_{i=1}^n\|\ab_i\|_2^2.
\end{align*}
\end{lemma}

\end{document}